\documentclass[review]{elsarticle}

\usepackage{lineno,hyperref}
\modulolinenumbers[5]

\journal{Knowledge-based systems}

\usepackage{xcolor}
\usepackage{subfigure}
\usepackage{multirow}
\usepackage{array}
\usepackage{bm}
\usepackage[inline]{enumitem}

\newcommand{\floor}[1]{\lfloor #1 \rfloor}
\newcommand\myeq{\mkern1.5mu{=}\mkern1.5mu}

\begin{document}

\begin{frontmatter}

\title{DeepVATS: Deep Visual Analytics for Time Series \tnoteref{mytitlenote}}
% \tnotetext[mytitlenote]{Fully documented templates are available in the elsarticle package on \href{http://www.ctan.org/tex-archive/macros/latex/contrib/elsarticle}{CTAN}.}

% \address{Radarweg 29, Amsterdam}
% \fntext[myfootnote]{Since 1880.}

\author[upmAddress]{Victor Rodriguez-Fernandez\corref{mycorrespondingauthor}}
\ead{victor.rfernandez@upm.es}
\cortext[mycorrespondingauthor]{Corresponding author}

\author[viuAddress]{David Montalvo}
\ead{david.montalvo@campusviu.es}

\author[uninaAddress]{Francesco Piccialli}
\ead{francesco.piccialli@unina.it}

\author[ujAddress]{Grzegorz J. Nalepa}
\ead{grzegorz.j.nalepa@uj.edu.pl}

\author[upmAddress]{David Camacho}
\ead{david.camacho@upm.es}

\address[upmAddress]{School of Computer Systems Engineering, Universidad Polit\'ecnica de Madrid, Calle de Alan Turing, 28038 Madrid, Spain}
\address[viuAddress]{Valencian International University, Calle Pintor Sorolla 21, 46002 Valencia, Spain}
\address[uninaAddress]{Department of Mathematics and Application ‘R. Caccioppoli’, University of Naples Federico II, Italy
}
\address[ujAddress]{Jagiellonian Human-Centered Artificial Intelligence Laboratory (JAHCAI), Institute of Applied Computer Science, Jagiellonian University, 30-348 Krakow, Poland}

\begin{abstract}
%TODO: Complete once we finish the experiments and conclusions
The field of Deep Visual Analytics (DVA) has recently arisen from the idea of developing Visual Interactive Systems supported by deep learning, in order to provide them with large-scale data processing capabilities and to unify their implementation across different data and domains. In this paper we present DeepVATS, an open-source tool that brings the field of DVA into time series data. DeepVATS trains, in a self-supervised way, a masked time series autoencoder that reconstructs patches of a time series, and projects the knowledge contained in the embeddings of that model in an interactive plot, from which time series patterns and anomalies emerge and can be easily spotted. The tool includes a back-end for data processing pipeline and model training, as well as a front-end with a interactive user interface. We report on results that validate the utility of DeepVATS, running experiments on both synthetic and real datasets. The code is publicly available on \url{https://github.com/vrodriguezf/deepvats}
\end{abstract}

\begin{keyword}
Deep learning \sep Visual Analytics  \sep Time series \sep Masked AutoEncoder
%\MSC[2010] 00-01\sep  99-00
\end{keyword}

\end{frontmatter}

%\linenumbers

\section{Introduction}

% \paragraph{Installation} If the document class \emph{elsarticle} is not available on your computer, you can download and install the system package \emph{texlive-publishers} (Linux) or install the \LaTeX\ package \emph{elsarticle} using the package manager of your \TeX\ installation, which is typically \TeX\ Live or Mik\TeX.

Deep learning is the field of Artificial Intelligence that studies the creation of learning systems through the use of artificial neural networks. In the last few years, this field has experienced greater growth than any other in computer science, becoming the quintessential tool for solving any type of task related to the areas of computer vision and natural language processing.

While the research and development of deep learning systems for image and text data is already very consolidated, there are other important data modalities for which these methods are yet to take off. Among these modalities, time series stand out, due to their ubiquitous presence in industrial, medical, financial processes, etc., and due to their availability on an unprecendented scale, which is caused by the sensorisation of the world in which we live (the so-called Internet of Things, or IoT) and the increase in storage capacity and massive data processing \cite{ismail2019deep}.

Nowadays, there is a demand for solutions that help data analysts to understand time series data, especially when these are of long duration, due to the information overload that they entail. Both the purely exploratory tasks, such as finding cyclical patterns, anomalies, and clusters, as well as the predictive ones (classification, forecasting...) are, in most cases, carried out by visual analytics systems built ad-hoc, whose computational basis is based on statistics and KPIs that are not transferable between domains, and that are not scalable to big datasets. To solve these complex issues, Deep Visual Analytics (DVA) is starting to grow as an alternative way of designing and implementing Visual Interactive Systems (VISs) powered by neural networks \cite{islam_deep_2021}. The reason why DVA systems, or more generally, learning-based analytics, are especially useful for long time series is that the model empowering the system does not have to be trained on the whole series, but just on a representative slice of it. Once the model is trained, it can be applied to the whole series, or even to future data coming from the series in a fast way, as long as there are not remarkable distribution shifts in the data.

\begin{figure}
    \centering
    \includegraphics[width=\textwidth]{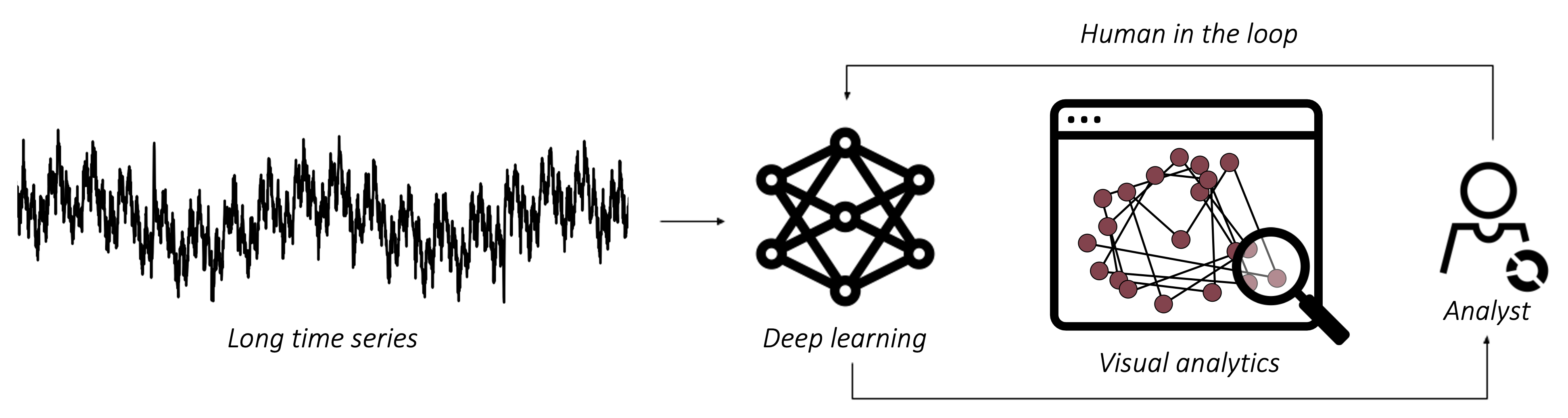}
    \caption{General outline of DeepVATS.}
    \label{fig:flowchart_concept}
\end{figure}

In this work, we present DeepVATS (Deep Visual Analytics for Time Series), an open-source tool \footnote{Link to Github repository: \url{https://github.com/vrodriguezf/deepvats} \label{fn:dvats-github}} that combines cutting-edge research in neural networks and visual analytics. It is inspired by projects such as \emph{TimeCluster} \cite{ali_timecluster_2019}, \emph{PSEUDo} \cite{yu_pseudo_2021}, the TensorFlow's embedding projector \cite{smilkov_embedding_2016}, the OpenAI microscope \cite{noauthor_openai_2020} or \emph{platform.ai} \cite{noauthor_codeless_nodate}, in which tools are created to interpret the latent space of trained neural networks. These tools have proven how the internal representations (a.k.a \emph{embeddings}) of a neural network reveal high-level abstraction patterns present in the data, such as semantic similarity between words in the case of networks trained with natural language. Therefore, given a time series dataset, DeepVATS will carry out three main tasks:
\begin{enumerate}
    \item Train neural networks to obtain data representations that contain, in a compressed way, meaningful patterns.
    \item Project the content of the latent space of the trained neural network in a way that allows the detection of clusters and anomalies.
    \item Provide interactive visualisations to explore different perspectives of the projected latent space.
\end{enumerate}

% Self-supervised learning as a novelty
In order to allow an analysis and exploration of the time series data without the need of external expert labels, neural network training will be carried out through self-supervised learning techniques \cite{liu2021self}. In this sense, DeepVATS innovates upon the related work by implementing a \emph{Masked Time Series AutoEncoder} (MTSAE), with the simple but powerful approach of randomly masking steps of the input series and try to reconstruct them during training. Unlike classic autoencoders such as the one used in \emph{TimeCluster}, we show that MTSAEs are more sample efficient, and provide more flexibility in the creation of latent spaces that are useful for visual analytics, due to the different possibilities in the design of the masking strategy. To the best of our knowledge, this is the first work in which latent spaces provided by MTSAEs are employed for visual analytics purposes. Furthermore, it is aligned with the recent trend in computer vision and self-supervised learning, which places Masked AutoEncoders as scalable and state-of-the-art vision learners \cite{He2021MaskedAA}.

The rest of the paper is structured as follows: Section \ref{sec:backgrounds} provides the background knowledge in each of the fields required to understand how our tool works, as well as related works that have inspired and influenced this work. In Section \ref{sec:deepvats}, we describe our system for deep visual analytics of time series, DeepVATS, detailing both the back-end, which includes the data processing pipeline and model training, and the front-end, which includes the user interface and an in-depth explanation of how to actually use the tool. Then, in Section \ref{sec:experiments}, we report on experiment results that validate the utility of our solutions, running experiments on both synthetic and real datasets, and comparing the results with other tools for deep visual analytics of time series. Finally, in Section \ref{sec:conclusions} the conclusions and outline future directions and research questions are given.

\section{Backgrounds and Related Work} \label{sec:backgrounds}
% VIS -> DVA -> DL & self-supervised learning -> Examples with image and text -> Examples with time series
% Platforms: Start with text and image (microscope, embeddings projector, wandb embeddings projector), and finish with time series (PSEUDO and Timecluster)
% Talk a bit of time series with the concept of TimeCluster
Visual Analytics (VA) can be defined as the study, methods and techniques that can be used to turning raw information into visual representations of data and knowledge. It refers to the use of interactive visual interfaces and statistical visualisations in the context of recognising emerging patterns and trends within the data, that show underlying relationships and support its understanding. It is an essential component of Decision Support Systems and large-scale decision making, especially in a business or commercial environment \cite{arnott2015critical}. Examples of complex problem areas solved by data visual analytics include cybersecurity, healthcare, chemistry, social science, astronomy, and physics, among others \cite{8740868}.

One of the main challenges of VA is the scalability, which involves both human and machine limitations \cite{8740868}. VA techniques need to be able to scale with both the size and the dimension of the data. Another challenge is the unification and standardization of VA approaches, since nowadays, different areas and different data modalities present VA solutions completely independent to each other, which hinders the knowledge transfer between systems, and, ultimately, decelerates the progress in the field.

% Deep learning as field that presents standardisation and 
The two challenges of VA introduced above are exactly two of the key qualities that can be found in the field of deep learning. Deep learning is a sub-field of Machine learning that uses multi-layer neural networks to extract patterns from data. Each of these layers, consisting of a set of artificial neurons, takes knowledge extracted from the previous layers and gradually refines it by applying both linear (matrix multiplication) and non-linear operations (activation functions such as ReLU) between them. Layers are laid out following a specific architecture, and are trained by gradient descent-based optimisation algorithms to minimise the values of a loss function. As long as large datasets are available, deep learning models have proven to be superior to classic machine learning algorithms in both accuracy and scalability \cite{Oerlemans2020SpecialIO}. Furthermore, the knowledge and expertise in deep learning can be transferred between domains and data modalities, which makes it a ``Swiss army knife" to implement any learning system.

% Deep visual analytics - embeddings and latent space
Leveraging the scalability and standardization of deep learning approaches can be helpful for VA. In this sense, the intersection between the two fields have evolved into Deep Visual Analytics (DVA) \cite{islam_deep_2021}. Normally, DVA systems apply deep learning to get predictive insights on the data, on top of other classic data visualisations. However, the whole exploration part of a VA solution can be addressed with deep learning too, by analysing the so-called ``latent space" of the neural network. This is an intermediate layer of the network that has learnt to encode its input data in an abstract and meaningful way for the training task at hand. These representations are often known as embeddings, i.e., mappings from one vector space (the input layer) to another vector space (the latent intermediate layer), normally with lower dimensionality.

% Autoencoders
Autoencoders are a particular type of deep learning architectures that are specially useful for exploring latent spaces. They are neural networks that have been designed to reproduce their input as output, without any explicit labels or supervision \cite{baldi_2021}. Despite the fact that autoencoders can be used for reconstruction tasks, the main purpose of these architectures is to learn a meaningful latent representation of the input data. This representation is usually learned in the bottleneck layer, which is the most compressed layer of the architecture. To reconstruct the input, the autoencoder needs to decode the latent representation back to the original input space. Therefore, by analysing the activations of the bottleneck layer, it is possible to gain insights about the input data.

% Example of embedding analysis (preferably supervised, to then link the self-supervised paragraph)
The analysis of neural latent spaces is a common practice in the field of eXplainable AI (XAI) \cite{arnott2015critical}. This field aims at improving the interpretability of complex predictive models such as deep neural networks, which are often criticised as being merely black boxes with good performance and success on predictive tasks, but poor adoption in applications where interpretability is essential \cite{DBLP:journals/corr/ShrikumarGSK16}. A popular technique for explaining neural networks is Layer-Wise Relevance Propagation (LRP), which operates by propagating the model prediction backward in the neural network. LRP has been successfully applied in interpretable computer vision applications, speech recognition, and predictive maintenance, among other areas \cite{Montavon2019LayerWiseRP}. This idea of exploring how neural networks work by analysing them at the level of layer activations was taken to the level of a usable application by the company OpenAI, with \emph{The OpenAI Microscope} \cite{noauthor_openai_2020}. This is a collection of visualisations of every significant layer and neuron of 13 important vision models trained on huge datasets.

The XAI techniques mentioned above use the dataset (input data and output labels/predictions) to explain a model, normally trained in a supervised way. However, in the case of DVA, we want a model to explain the dataset. An analyst may want to explore a raw dataset without giving any additional information that is used to train a supervised model, or without even knowing that there is a deep learning model behind. In this sense, self-supervised learning is an exciting research direction that aims at learning data representations using labels that are embedded in the data itself, without explicit and potentially even manual supervision. Many of these methods are developed in specific communities such as natural language processing, computer vision or graph learning \cite{liu2021self}. One of the major benefits of self-supervised learning is the ability to scale to large amounts of unlabelled data in a lifelong learning manner.

Although there are many techniques for pre-training models in a self-supervised way, in this work we draw inspiration from the Masked Language Model (MLM) methods used originally in Natural Language Processing (NLP). Here, the input sentences from a text dataset are randomly masked, and the aim of the model is to recover the masked word. This technique gave rise to BERT \cite{devlin_bert_2019}, a breakthrough language model that was pre-trained using an enormous dataset, and that could be applied to a variety of different downstream tasks such as sentiment analysis. More recently, He et. al in \cite{He2021MaskedAA} showed that this same concept can be applied in computer visions, and popularised the concept of Masked Autoencoders (MAEs), as scalable self-supervised learners of large visual representations, outperforming other approaches. In time series, Zerveas et. al presented in \cite{10.1145/3447548.3467401} a presented a novel framework for BERT-like self-supervised representation learning of multivariate time series using the transformer architecture, which will be used as the basis for the Masked Time Series AutoEncoder (MTSAE) employed in this work.

% related work (platform. ai -> embeddings projector ->
There are many examples of DVA systems supported by a neural network trained in a self-supervised way. In computer vision, \emph{platform.ai} \cite{noauthor_codeless_nodate} stands out as a human-in-the-loop development environment to train computer vision models without the need of knowledge of deep learning or code skills. A backbone self-supervised AI model is used to help domain experts identify patterns in the data by displaying visual groupings or clusters that can easily be labelled. Expert labels are then used to fine-tune the backbone model, which improves the clusters presented to the user. This process is repeated iteratively creating an active learning framework that includes both human and machine in the learning process. 

% NLP -> embeddings projector
In natural language processing, Google's embedding projector \cite{smilkov_embedding_2016} is a clear example of what a DVA tool can do with the embeddings of a self-supervised language model, such as BERT, and it is a direct inspiration for the tool described in this manuscript. They propose and implement a toolkit for interactively viewing and analysing textual embeddings and other unstructured data, projecting them in a 2D or 3D space via dimensionality reduction algorithms, such as T-SNE \cite{van2008visualizing} or UMAP \cite{mcinnes2018umap}. With such a system, one can clearly see how words with semantic similarity appear together in the embedding space.

% Time series - PsEUDO 
Finally, in the case of time series data, the reception of deep learning as the de facto technique to implement learning systems has not been as fast and significant as in language and vision, due to the accuracy improvements for common tasks such as time series classification has not been that drastic \cite{Jiang2020TimeSC}. Still, in recent years, we are seeing how the latest advances in deep learning, such as self-supervised strategies and transformer-based architectures, are being transferred to time series data too \cite{Lim2021TimeseriesFW}.

In the field VA, there are several AI-supported solutions, and in fact, recent surveys have been published that explore the intersection between AI and data visualisation \cite{9523770, 9495259}. For time series data, we start by enumerating some VA systems that don't use deep learning as a backbone, such as:
\begin{enumerate*}
\item \emph{PSEUDo} \cite{yu_pseudo_2021}, which offers a novel adaptive feature learning technique for exploring visual patterns in multivariate time series. 
\item \emph{MTV} \cite{10.1145/3512950}, a visual analytics system aimed at supporting human-AI collaboration in the detection, investigation, and annotation of time series anomalies
\item \cite{https://doi.org/10.1111/cgf.14498}, which presents a linked-view visual analytics application for analysing high-dimensional measurement data with varying sampling rates in intensive care units.
\item \emph{Sintel} \cite{alnegheimish2022sintel}, a machine learning framework for end-to-end time series tasks such as anomaly detection.
\item \emph{MultiDR} \cite{9216630}, a dimensionality reduction framework that enables processing of multivariate time series as a whole to provide a comprehensive overview of the data.
\item \emph{Plotly Resampler} \cite{Fujiwara_2021}, an open source Python toolkit to effectively visualise large volumes of high frequency time series data.
\item \emph{TimeSeer} \cite{6200267}, a method and application for organising and exploring multivariate time series data using the so called \emph{Scagnostics} technique, which characterises the 2D distributions of orthogonal pairwise projections on a set of points in multidimensional Euclidean space.
\item \emph{TimeCurves} \cite{bach2015time}, a general approach for visualising patterns of evolution in temporal data, which can be applied to any dataset where a similarity metric between temporal snapshots can be defined. Visually, the curves look similar to the ones of this work, though the process to create them is based on Multi-Dimensional Scaling (MDS) \cite{torgerson1952multidimensional}
\item \cite{7192717} presents a new approach for exploring dynamic networks, which  enables the identification of stable and recurring states in the network. Although it is focused on dynamic networks and not time series in general, the resulting visualisations and the procedure presents analogies with the methodology of DeepVATS.
\end{enumerate*}

% DVA and time series - Timecluster
With regard to the specific field in which this work operates, i.e., DVA for time series, we can highlight the following related works:
\begin{enumerate}
    \item \emph{TimeCluster}, by Ali. et al \cite{ali_timecluster_2019}, which is arguably the closest related work to this one. TimeCluster proposes an interactive visual analytics system for exploring clusters (segments) in long time series in a single image displayed as a connected scatter plot. The input series is first processed with a sliding window approach, creating a dataset of smaller subsequent time windows that is then used to train a Deep Convolutional AutoEncoder that reconstructs the full windows. Then, the latent space of the model will be extracted and projected into the scattered plot.
    
    \item \emph{Peax}, by Lekschas et al. \cite{lekschas2020peax}, which uses a convolutional autoencoder for interactive visual pattern search in sequential data, adapting to user perception and improving pattern retrieval. Although it features a latent visible latent space to interact with, the main power of Peax lies on the idea of active learning to polish the results based on the user feedback.

    \item \cite{Guo2020ComparativeVA} introduces a method for calculating similarity between medical records using event and sequence embeddings derived from an autoencoder, along with a visual analytics system to support comparative studies of patient records.

    \item \emph{V-Awake} \cite{https://doi.org/10.1111/cgf.13667}, a visual analytics approach for correcting faulty predictions in sleep staging coming from a deep learning model. Although a deep learning model is used to get the predictions, the system does not include a visualisation of the latent space of that model.

    In \cite{9119141}, the authors propose a prototype of a sketch-based querying visual system for time series data that allows users to explore that data interactively without the need to set any parameter, just drawing a sketch. They used a pair of LSTM networks with shared parameters to encode the sketch and the time series data.
\end{enumerate}

The system proposed in this work, DeepVATS, is built on top of the ideas of some of the above-mentioned works, and inherits some of their data processing pipelines and graphical interface components. However, DeepVATS expands the ideas of these works in multiple points:
\begin{itemize}
    \item The backbone deep learning model of DeepVATS is a Masked Time Series AutoEncoder, in contrast to the classic autoencoders built in the other works such as TimeCluster or Peax, that reconstructs a full time window. This allows DeepVATS to be used in other tasks than segmentation, such as outlier detection, which is hard to accomplish by a normal autoencoder, given that they naturally tend to denoise the input signal and omit outliers when these are peaks.
    
    \item The DeepVATS backbone model is trained with a multi-window size strategy, so that the embeddings of the the trained model is less sensitive to the choice of window size, which is usually a critical value in the analysis of long time series.
    
    \item DeepVATS is available open-source to be used by any stakeholder in any appropriate dataset. Furthermore, it has been developed in a modular way with a general purpose, such that it is easy to replace any of the components (e.g., the backbone model).
\end{itemize}

\section{DeepVATS system description}
\label{sec:deepvats}
% architecture - nbdev, tsai, wandb, docker

\begin{figure}[h!]
    \centering
    \includegraphics[width=\textwidth]{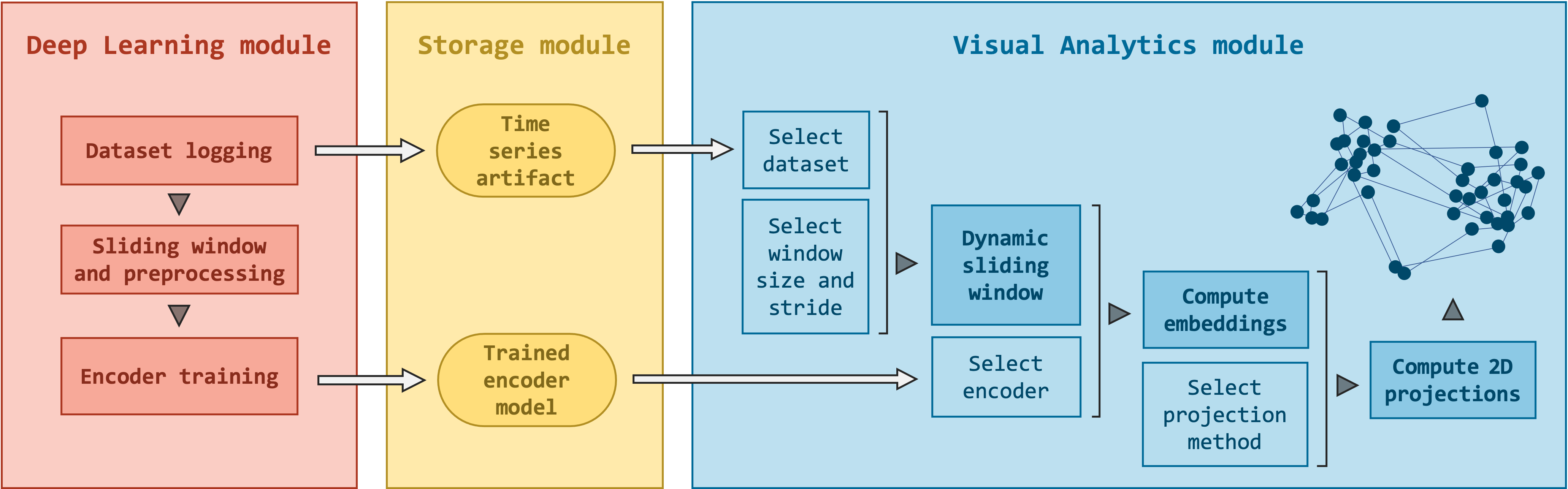}
    \caption{DeepVATS software architecture by modules.}
    \label{fig:architecture}
\end{figure}

This section describes the DeepVATS tool in detail, from the pipeline established to train the ``backbone" neural network model, to which we will refer to as the DL module, to the usage of the trained model in an exploratory way through a Graphical User Interface (GUI), namely the Visual Analytics (VA) module. An overview of the DeepVATS software architecture can be seen in Figure \ref{fig:architecture}. The DL module has been implemented as a Python library developed with \emph{nbdev} \cite{fastainb83:online}, a tool to create Python modules with Jupyter Notebooks. To train the models we have used the library \emph{tsai} \cite{tsai}, a state-of-the art deep learning library for time series in Pytorch. On the other hand, the VA module has been implemented using \emph{Shiny} \cite{shiny}, a popular library to build interactive web-apps in R.

% middleware wandb - storage module
A third module, namely the storage module, is placed between the two modules introduced above. It provides an API that allows to save the datasets and encoder models produced by the DL module, and load them into the VA module to be used for inference. For the implementation of this module we have relied on \emph{Weights \& Biases} \cite{wandb}, a tool for doing experiment tracking in machine learning projects. More specifically, Weights \& Biases provides a so-called \emph{Artifacts API} to save and version datasets and machine learning models as \emph{artifacts}, acting as a high-level domain specific database for machine learning.

% docker
DeepVATS has been created with the purpose of expanding the community of DVA for time series. For that reason, the tool is publicly available in a Github repository \ref{fn:dvats-github}, with Docker integration to make its deployment easier. Software-wise, some stages of the pipeline have been designed in a decoupled plug-and-play way, so that future researchers and data scientists can extend it to their needs.

\subsection{Deep Learning (DL) module}
In the Deep Learning (DL) module, an input time series dataset is loaded, processed and used to train a deep neural network encoder whose latent space will be analysed and used as the basis of the VA module. Two stages comprise the pipeline of this module, namely the dataset logging and the encoder training.

\subsubsection{Dataset logging} \label{subsubsec:dataset_logging}
Datasets have to be ``logged" as an artifact in the storage module before they can be used in any other part of the system. In order to define a time series dataset to be used in DeepVATS, some considerations have to be taken:
\begin{itemize}
    \item Only one time series at a time can be analysed. Therefore, the tool is suitable for long time series that present cyclical patterns and anomalies, and not in cases when the time series is not cyclical or when one wants to detect similarities differences between different samples (e.g., clinical records of different patients). In the future, we are planning to extend the tool so that it is applicable to datasets made of multiple time series too.
    
    \item The tool works in the same way regardless of the number of variables present in the time series, i.e., whether the time series is univariate or multivariate. Additionally, natural timestamps (i.e., datetimes) are not needed, since the time information is not used as input in the neural network encoder.
    
    \item The series is assumed to be regular, i.e., time steps must be evenly spaced.
    
    \item Missing values in the series, if any, are imputed through linear interpolation.
    
    \item In case the time series is too long, the user can choose an option to resample it into a lower frequency. The data will then be aggregated using the mean. 
    
\end{itemize}

\begin{figure}[h!]
    \centering
    \includegraphics[width=0.9\linewidth]{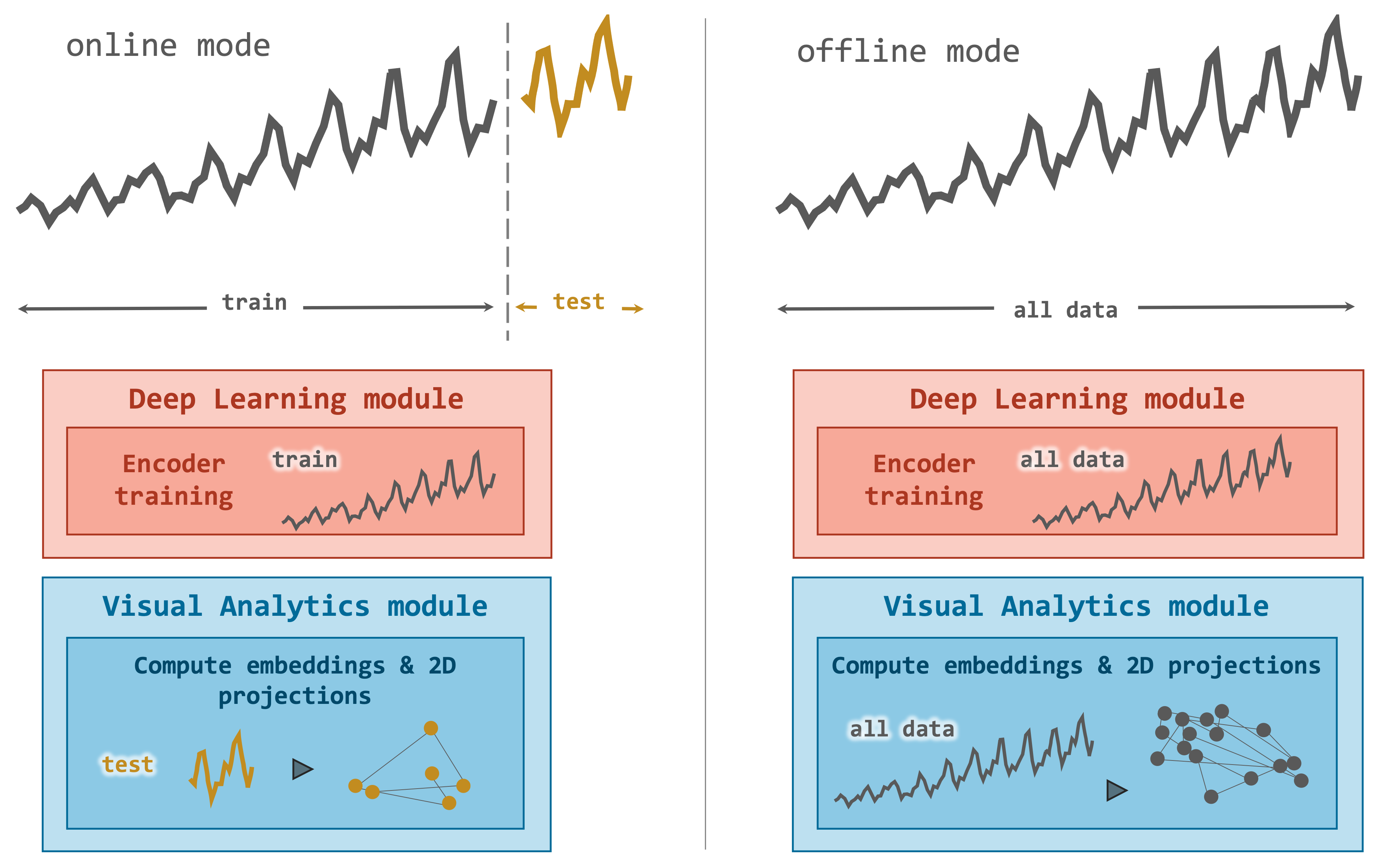}
    \caption{Online vs. offline mode in DeepVATS.}
    \label{fig:offline_online}
\end{figure}

% and the validation set needed to watch for model overfitting will be created by holding out random time windows after the sliding window process (more details in the next section)
By default, the time series will be logged in the storage module as a single artifact. This is appropriate for enabling offline data analysis, where the range of data that is analysed is the same that has been used to train the backbone encoder. However, DeepVATS also provides the option of splitting the time series manually and logging multiple artifacts out of it, namely the training and test artifacts. This is especially relevant to examine the capabilities of DeepVATS for online analysis, i.e., for  situations where an expert needs to analyse new data from a process often, and therefore, retraining the backbone encoder every time new data is available is infeasible. In these cases, one has to test the trained model on future time windows (See Figure \ref{fig:offline_online}), to check if it produces satisfactory results out of the training interval. Note that this opens the possibility of doing zero-shot DVA, i.e., of having an instant DL-based analysis of a dataset without having to retrain or fine-tune the backbone model. Additionally, the online mode is also suitable for very long time series, where the cost of training the backbone model on the whole series is unnecessarily high, and just using a slice of data is enough to create a model that will then be used to process (in inference time) the whole series in a fast way.

\subsubsection{Encoder training} \label{subsubsec:encoder}
%TODO vrodriguezf: Relate the concept of time window with the word "patch", as it is done in the abstract
The encoder is the central piece of the DL module, and arguably of the whole DeepVATS framework. We also refer to it in this work as the backbone model, or simply the model. An encoder, in deep learning, is an artificial neural network used to turn raw input variables from a high-dimensional space into efficient and compressed vector representations of a lower dimensionality, often known as embeddings, hidden states, or just the model's latent space. These representations are typically used as an intermediate step in machine learning tasks.

% Sliding window
Training the encoder properly needs a dataset with a big number of time series samples. However, as seen in the previous section, the input artifact logged in the storage module is a single long time series. Therefore, before passing it to the encoder, the input series is divided into separate, contiguous and fixed-size windows of data, or \emph{patches}, in a process often known as sliding window. The window size in this process, to which we will refer as $w$, is then a crucial hyperparameter that can affect the dataset that is passed to the model, and thus, it can affect the representations obtained by the encoder and the visualisations that we get out of it. By default, the stride of the sliding window, i.e., the number of data points that the window is moved ahead along the series is set to 1. Therefore, given a time series of length $T$, a window size $w$ and a stride of $s$, the dataset after the sliding window will have a total of $N=\floor{\frac{T-w}{s}} + 1$ time windows. Despite using such a low stride causes contiguous time windows to almost completely overlap with each other, this prevents the training process from having a bias from the choice of this value.

% splitting & normalisation
In order to watch for overfitting, the dataset created with the sliding window process is split into training and validation subsets. We do this differently depending on whether there are one or two time series artifacts logged for the dataset in the storage module. If there are two, namely the training and test artifacts, we create the validation set by holding out the last 20\% of the training set, to be temporally coherent with respect to the fact that the model will be tested on future data. Otherwise, the validation set is created by randomly holding out 20\% of the time windows. With regard to normalisation, it can be applied either batch-wise, sample-wise, or dataset-wise, depending on the needs of the use case.

% MTSAE
The use of encoders in deep learning is commonly associated with a so-called encoder-decoder architecture, where there is a decoder network fed with the outputs of the encoder, which turns them into an actual prediction. One special case of the encoder-decoder architecture, which will be employed in this work, is the AutoEncoder (AE), in which the input and output spaces are the same, i.e., the task of the model is to reconstruct the input data, by learning first a compressed representation of it. Note that this is a self-supervised task, since the target needed to train the model is inherent in the data itself. Note also that, despite DeepVATS uses by default an autoencoder architecture, its modular design allows it to plug in any other encoder.

%\begin{figure}
%    \centering
%    \includegraphics[width=\linewidth]{figures/new_masking_strategies.png}
%    \caption{Different masking strategies available in DeepVATS.}
%    \label{fig:masking_strategies}
%\end{figure}

\begin{figure}[h!]
\centering
\subfigure[ ]{
\includegraphics[width=0.47\textwidth]{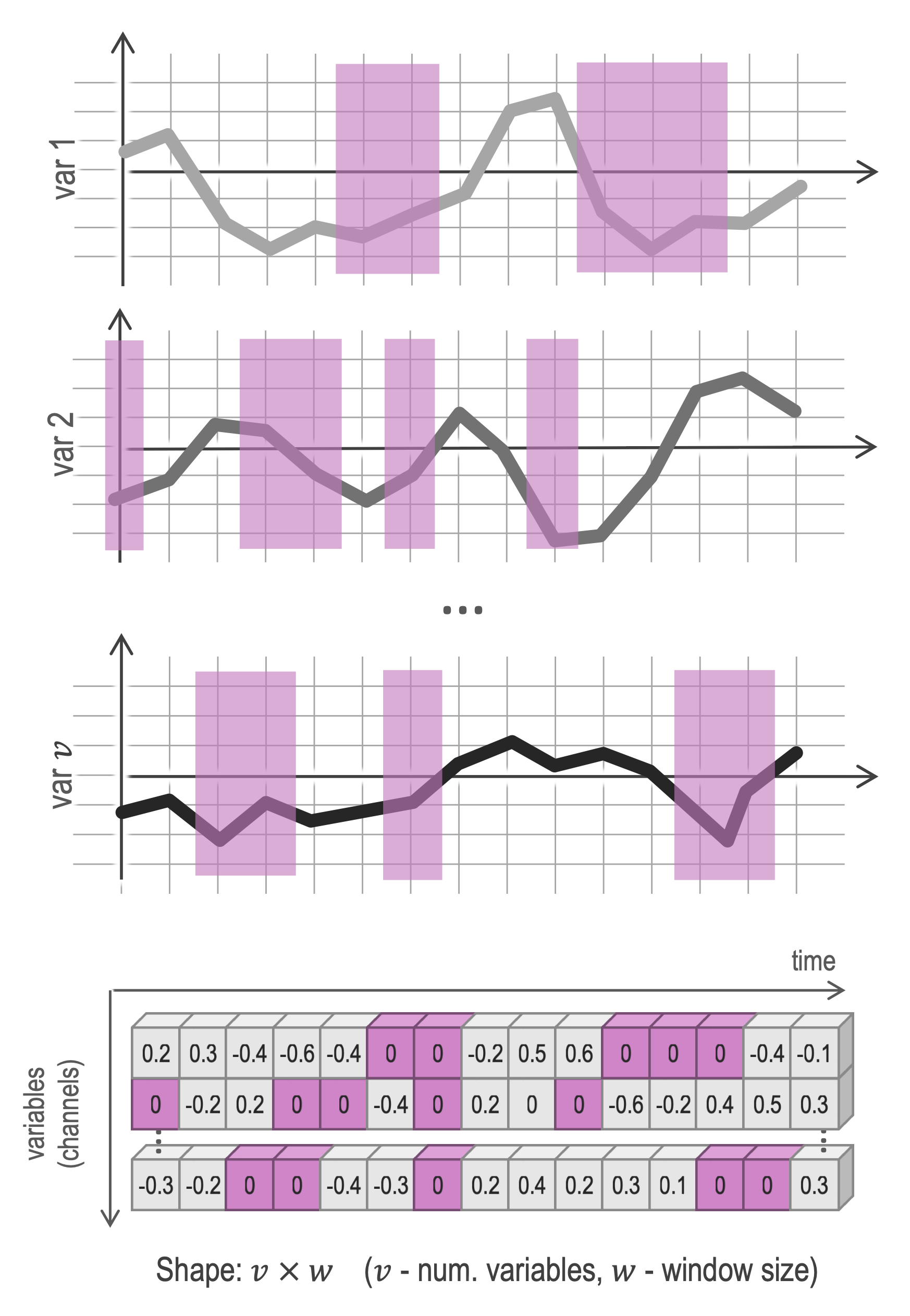}
\label{fig:masking_process}
}
\hfill
\subfigure[ ]{
\includegraphics[width=0.47\textwidth]{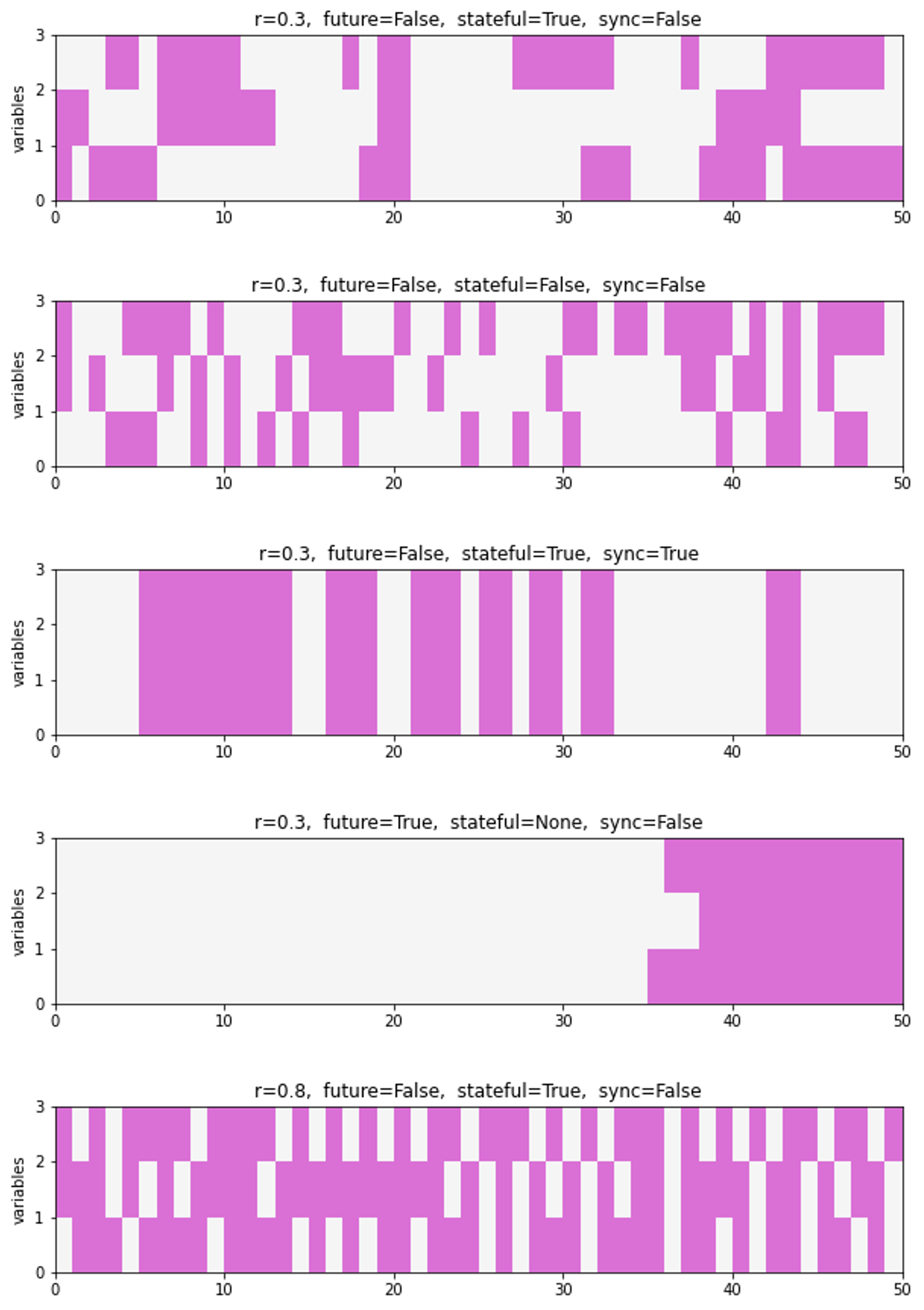}
\label{fig:masking_strategies}
}
\caption{Time series masking. (a) Graphical representation of the time series vectorization process, including the application of masks. (b) Different masking strategies available in DeepVATS.}
\end{figure}

% [Commented] Under this framework, a classic autoencoder is just a special case of a masked autoencoder, in which the whole time window is masked.
We refer to the autoencoder used in DeepVATS as the Masked Time Series AutoEncoder (MTSAE). The term ``mask" refers to the idea of hiding (or zeroing) sections of the time series, and training the model to reconstruct them \cite{10.1145/3447548.3467401}. One of the advantages of having a masked learning framework is the flexibility in the design of the masking strategy, as it can be seen in Figure \ref{fig:masking_strategies}. Masks are applied stochastically over the time series, controlled by the probability of masking, $r$. An uniform distribution is followed by default, but a ``stateful" mode can be selected too, in which a geometric distribution is applied so that average mask length meets the value of another parameter called $lm$. This is required in order to force the model beyond trivial prediction methods such as padding, replication or linear interpolation, which might offer a sufficiently good approximation for very short masked sequences. Additionally, one can elect to use the same mask synchronised over all variables for a given object (synchronised mode), or instead, generate a new mask for each in order to encourage the model to learn both relationships between different values along individual sequences, as well as inter-dependencies between variables in order to improve the modelling. If masks are applied at the end of the time series, the MTSAE acts effectively as a forecasting model, which makes forecasting a special case of a masked autoencoder. In any of these cases, a binary sequence is generated based on each variable in the series, and the model is tasked with predicting the values of the time series over which the mask is set to 0.

\begin{figure}[h!]
    \centering
    \includegraphics[width=0.9\linewidth]{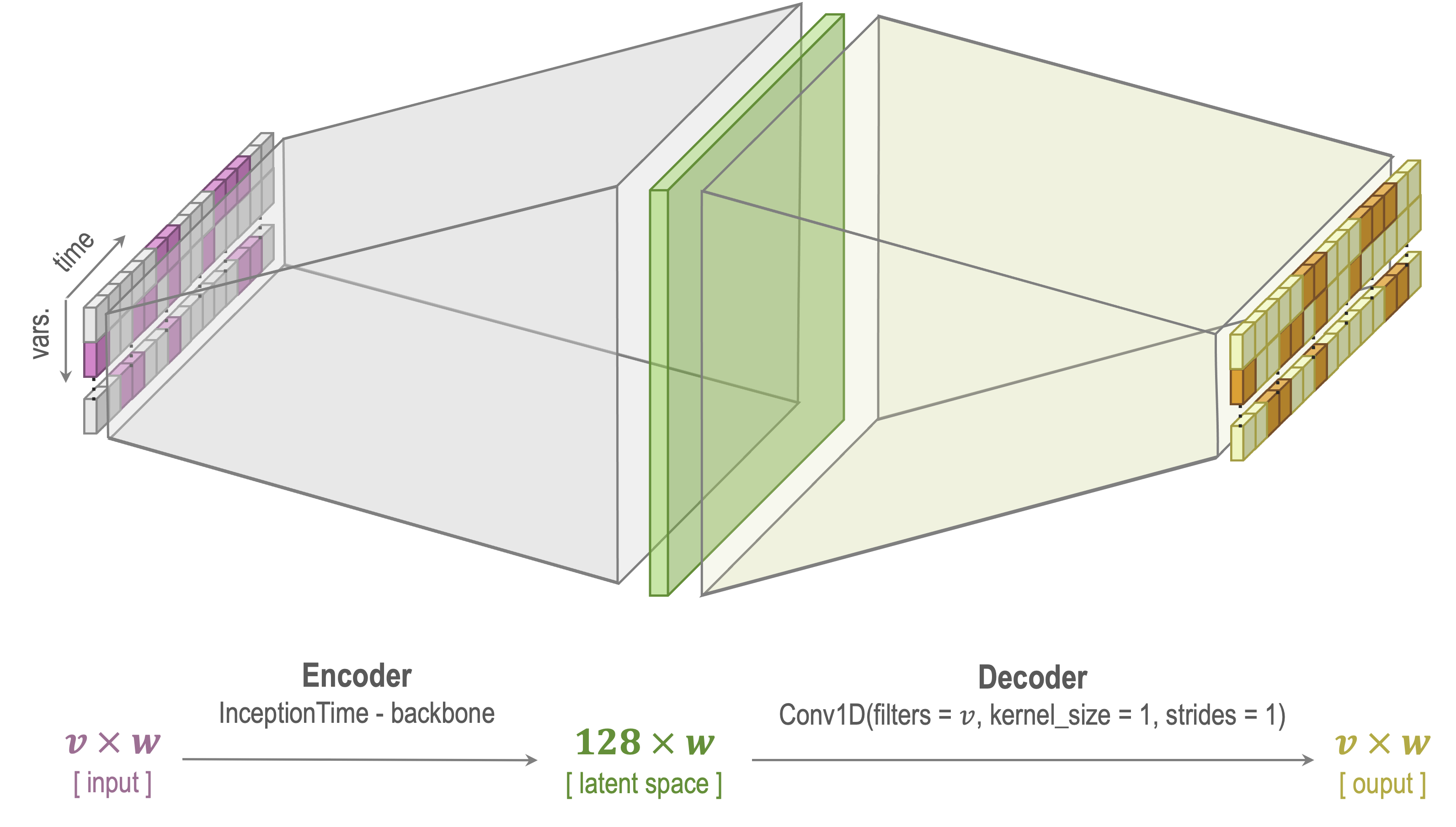}
    \caption{The Masked Time Series Autoencoder is based on the InceptionTime architecture, a popular 1D Convolutional Neural Network.}
    \label{fig:mtsae}
\end{figure}

% Encoder architecture - InceptionTime, but anything can be plugged thanks to tsai's MVP (Cite Zerveas). If a new encoder is added, then the last layer is considered the source of the visualisations. This is different to TimeCluster, where the autoencoder is a symetric DCAE.
With regard to the architecture used to build the MTSAE, we clearly differentiate the encoder and decoder sections. For the encoder, we employ the InceptionTime architecture \cite{fawaz_inceptiontime_2020}. This is, to the best of our knowledge, the best deep neural architecture for time series tasks among the family of 1-dimensional Convolutional Neural Networks (CNNs). In this work, we use a network made up of six sequential inception modules which maintain residual connections, a common strategy to build deeper CNNs. On the other hand, the decoder is made up of just one output layer, in which the shape of the original input data is reconstructed by using a convolutional layer with a number of filters equal to the number of channels in the input data, and a filter size of 1 (See Figure \ref{fig:mtsae}). Note that this decoder layer can be created in the same way regardless of the encoder architecture. Therefore, new encoder architectures can be plugged into the framework without making any change in the training process.

\begin{figure}[h!]
    \centering
    \includegraphics[width=\linewidth]{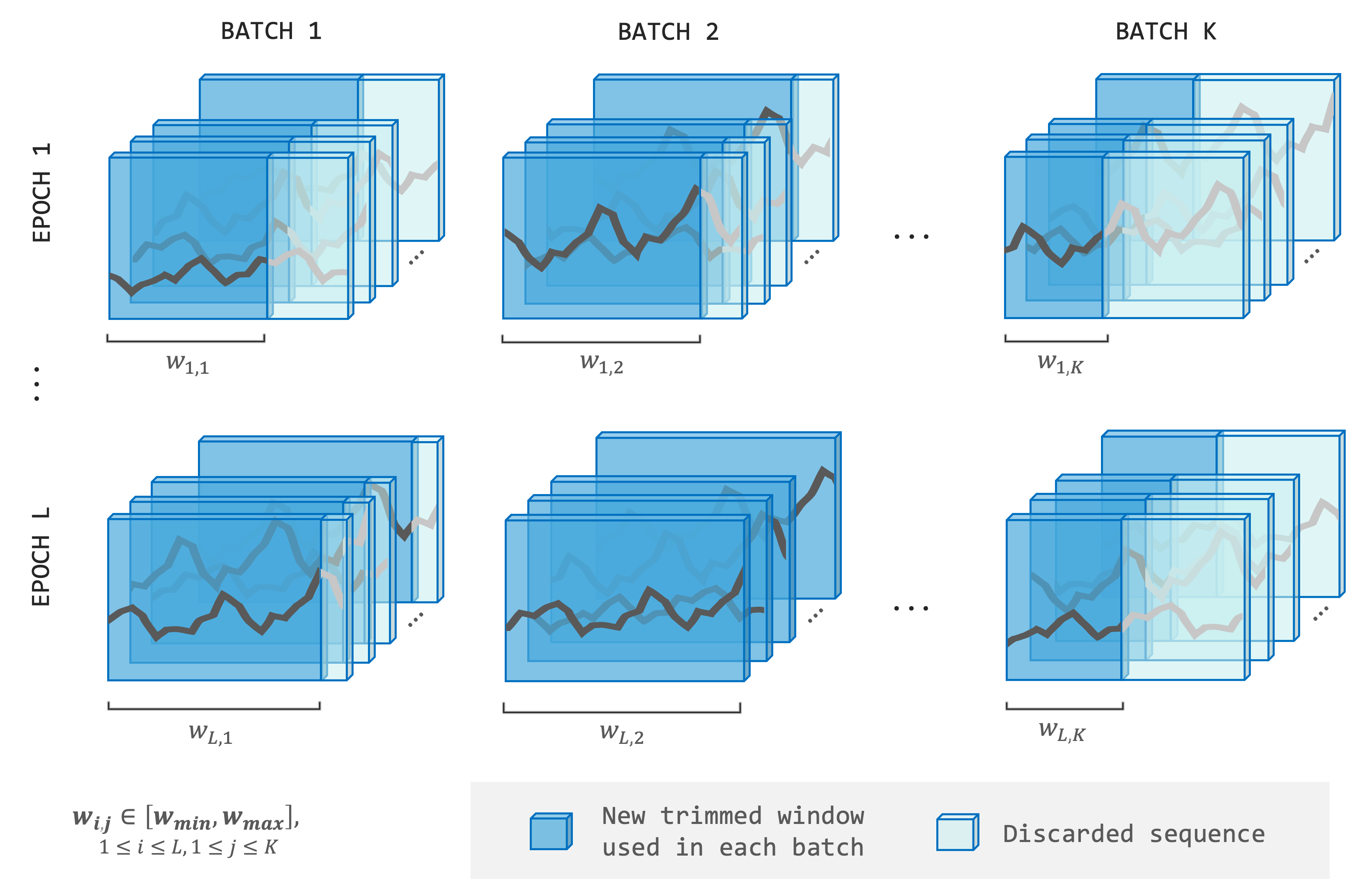}
    \caption{Training with variable window sizes trims the sequence length of each batch dynamically, in order to increase regularisation and robustness of the trained model.}
    \label{fig:variable_window_sizes}
\end{figure}

% Variable window sizes
One advantage of using a CNN such as InceptionTime in the encoder is that the neural architecture does not change depending on the sequence length of the input, unlike for vanilla RNNs or Transformers. This fact can be used to implement variable window sizes during training (See Figure \ref{fig:variable_window_sizes}), which allows the model to be trained with different window sizes in each training iteration, obtaining a model less sensitive to the choice of this hyperparameter. To do this, we set an interval $[w_{\textnormal{min}}, w_{\textnormal{max}}]$, and for each training iteration, the batch will be randomly truncated at a given value within that range. This has a regularisation effect too, which prevents overfitting in cases where the dataset is small. Note that, under this setup, $w_\textnormal{max}$ has to be lower or equals to the value of the window size $w$ used in the sliding window process. However, setting $w_{\textnormal{max}} < w$ causes that the data points between $w_{\textnormal{max}}$ and $w$ in each batch are lost, since they will never be passed to the model. Therefore, by default DeepVATS sets $w=w_\textnormal{max}$.

% Loss function
We train the model using an adaptation of the Mean Squared Error (MSE) metric, such that it only considers the predictions over masked values. For a given object with $i$ channels and $t$ time steps, this loss is defined as,

\begin{equation}
    \mathcal{L} = \mathop{\sum \sum}_{(t, i) \in M} (\hat{x}(t, i) - x(t, i))^{2},
    \label{eq: loss_func}
\end{equation}
% _{\text{MSE}}

\vspace{3mm}
\noindent where $x$ are the true values of the time series and $\hat{x}$ those predicted by the model over the set $M \equiv \{(t,i) : m_{t,i} = 0\}$, where $m_{t,i}$ are the elements of the mask $M$.

Once the model has been trained and validated properly, it is logged as an "encoder artifact" in the storage module so that it can be used for inference in the VA module. This artifact will contain not only the weights of the neural network, but also metadata associated to the training process such as the window size employed to slice the dataset, the masking type and probability, etc. 

\subsection{Visual Analytics (VA) module}

% @dmontalvo: Revisar captura para utilizar un modelo de los mostrados en experimentación
\begin{figure}[h!]
    \centering
    \includegraphics[width=\linewidth]{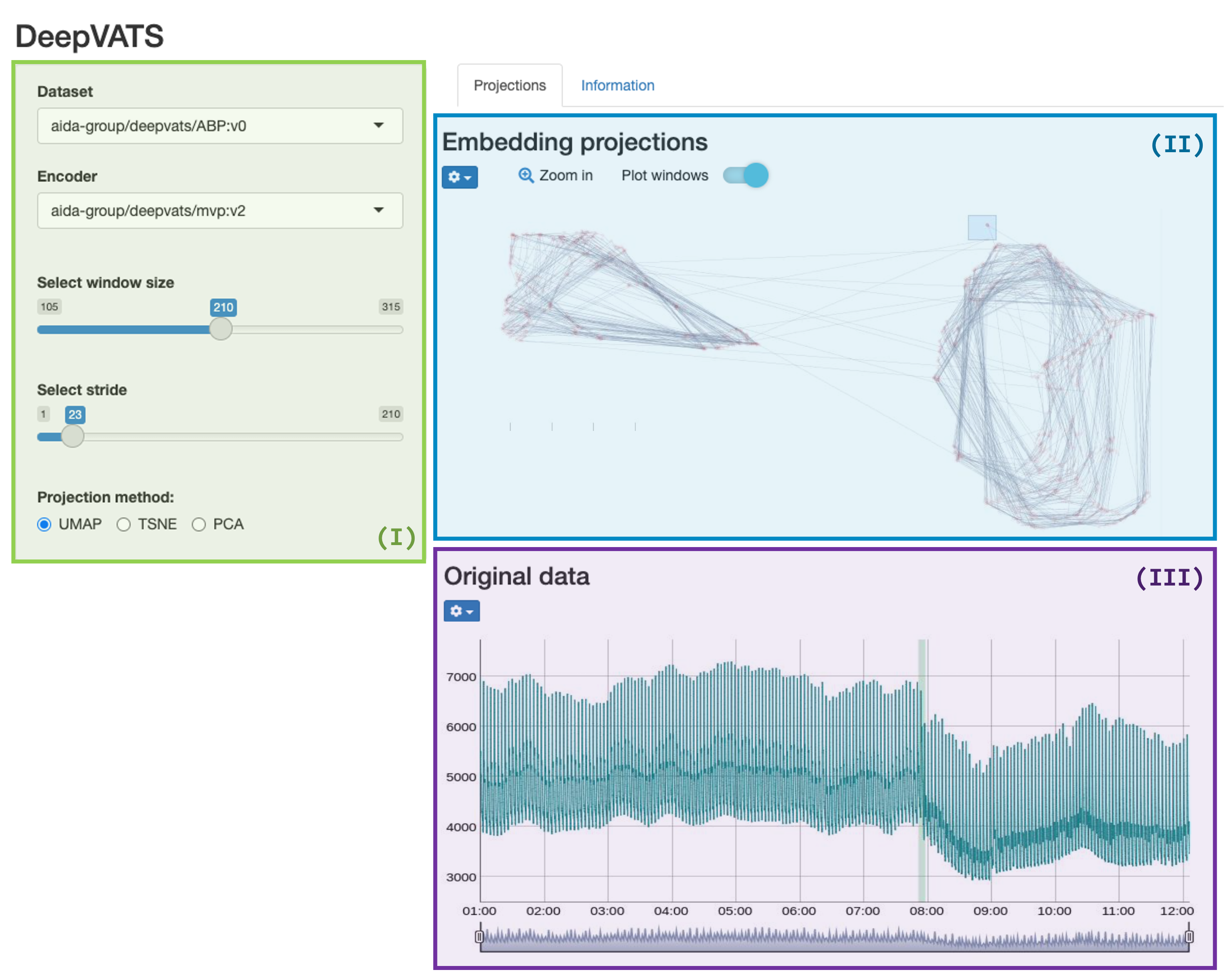}
    \caption{Graphical User Interface of DeepVATS. Note that there's a section of the embedding projections selected, which is in turn highlighted in the time series plot, showing the two-way communication between the plots.}
    \label{fig:gui}
\end{figure}

In the VA module, the embeddings of a time series dataset are computed using one encoder trained following the pipeline described in the previous section. Those embeddings are projected using dimensionality reduction techniques, and visualised as an interactive connected scatter plot in tandem with an interactive time series plot, whose interactions affect the view of each other. An overview of the GUI displayed by this module is shown in Figure \ref{fig:gui}. This has been divided into 3 sections, namely the control panel (Mark 1 in the figure), the embeddings projection (Mark 2), which contains the central connected scatter plot, and the time series plot (Mark 3).

\subsubsection{Control panel}
In this part of the GUI, the user can configure a variety of controls that will affect how the rest of the plots of the GUI look like. These controls are:
\begin{itemize}
    \item \textit{Dataset selector}: It allows the user to select the dataset that wants to be analysed, among the list of time series artifacts that have been logged before in the storage module by the DL module (See Section \ref{subsubsec:dataset_logging}).
    
    \item \textit{Encoder selector}: It allows the user to select the encoder that wants to be used to encode the selected dataset, among the list of encoders that have been trained in the DL module for the selected dataset and logged in the storage module (See Section \ref{subsubsec:encoder}).
    
    \item \textit{Window size selector}: The value selected here will be used to slice the selected time series dataset following a sliding window process. If the selected encoder has been trained with variable window sizes, this selector will allow to choose values in the interval $[w_{\textnormal{min}}, w_{\textnormal{max}}]$, i.e., the same one used for training the encoder. Otherwise, in case a fixed window size approach has been employed, the range of selectable values is $[w - w/2, w + w/2]$, where $w$ is the value of the window size used in training. In any case, by default the selector is set to the value $w$.
    
    \item \textit{Stride selector}: Unlike in training, where the stride of the sliding window process is always set to 1 to maximise the amount of data windows created, increasing this stride in inference can help create clearer projections of the embeddings which reveal better information. Additionally, it reduces the amount of data windows, reducing the inference time too, which is crucial for a fluid exploration of the data. The values to select for the stride range between 1 and the value of the window size chosen in the above selector.
    
    \item \textit{Projection method selector}: Three possible dimensionality reduction algorithms can be applied to the encoded data windows, namely Uniform Manifold Approximation and Projection (UMAP) \cite{mcinnes2018umap}, t-Distributed Stochastic Neighbor Embedding (t-SNE) \cite{van2008visualizing}, and Principal Component Analysis (PCA) \cite{abdi2010principal}. All of them are run directly on GPU using the RAPIDS library \cite{rapidsai}, in order to maximise their performance.
    
\end{itemize}

\subsubsection{Embedding projections}

%\begin{figure}
%    \centering
%    \includegraphics[width=\linewidth]{figures/compute_embeddings.png}
%    \caption{The inference process takes the activations from the last convolutional layer of the InceptionTime architecture and averages them on the sequence length.}
%    \label{fig:compute_embeddings}
%\end{figure}

\begin{figure}[h!]
\centering
\subfigure[ ]{
\includegraphics[width=0.52\textwidth]{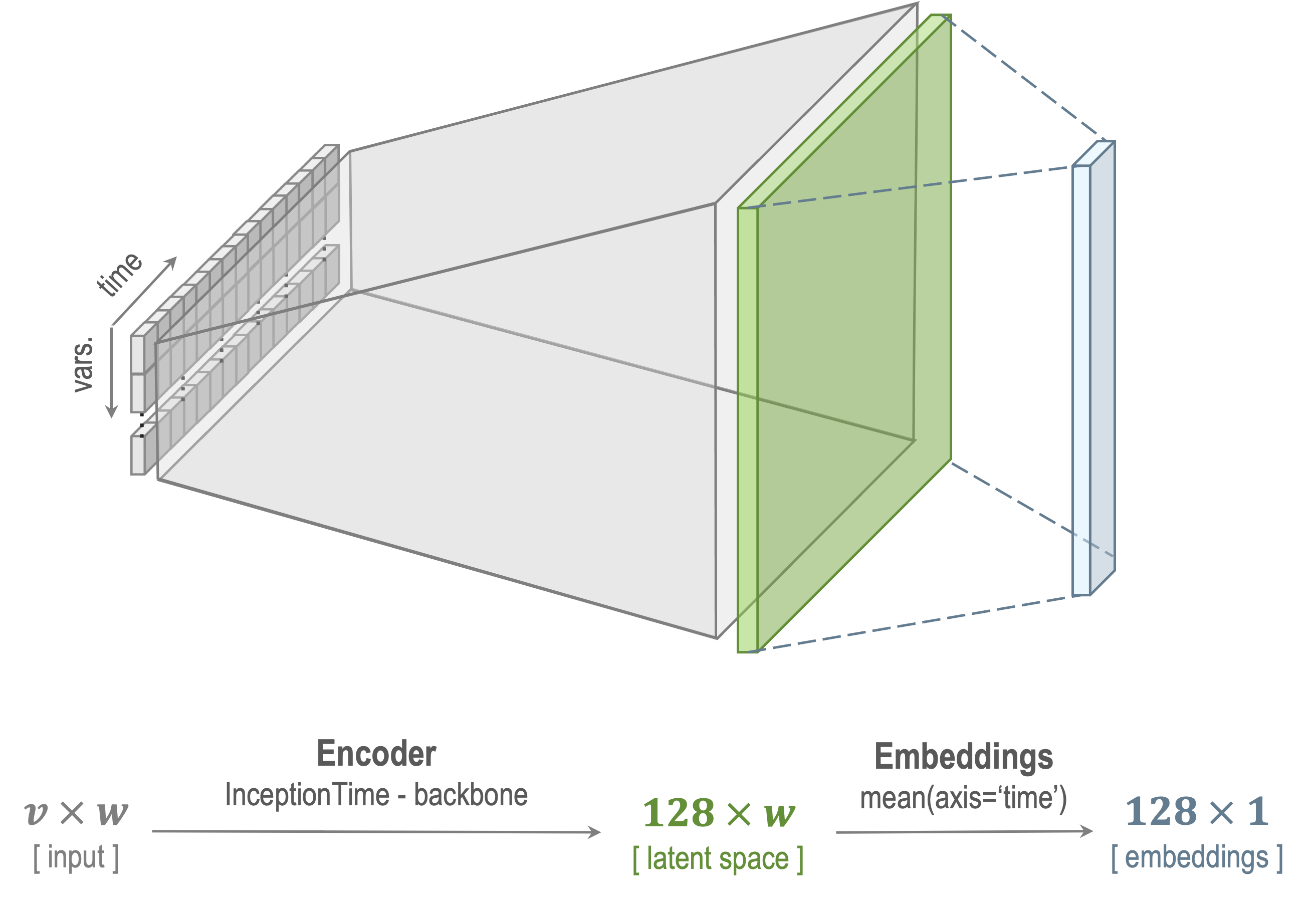}
\label{fig:compute_embeddings}
}
\hfill
\subfigure[ ]{
\includegraphics[width=0.38\textwidth]{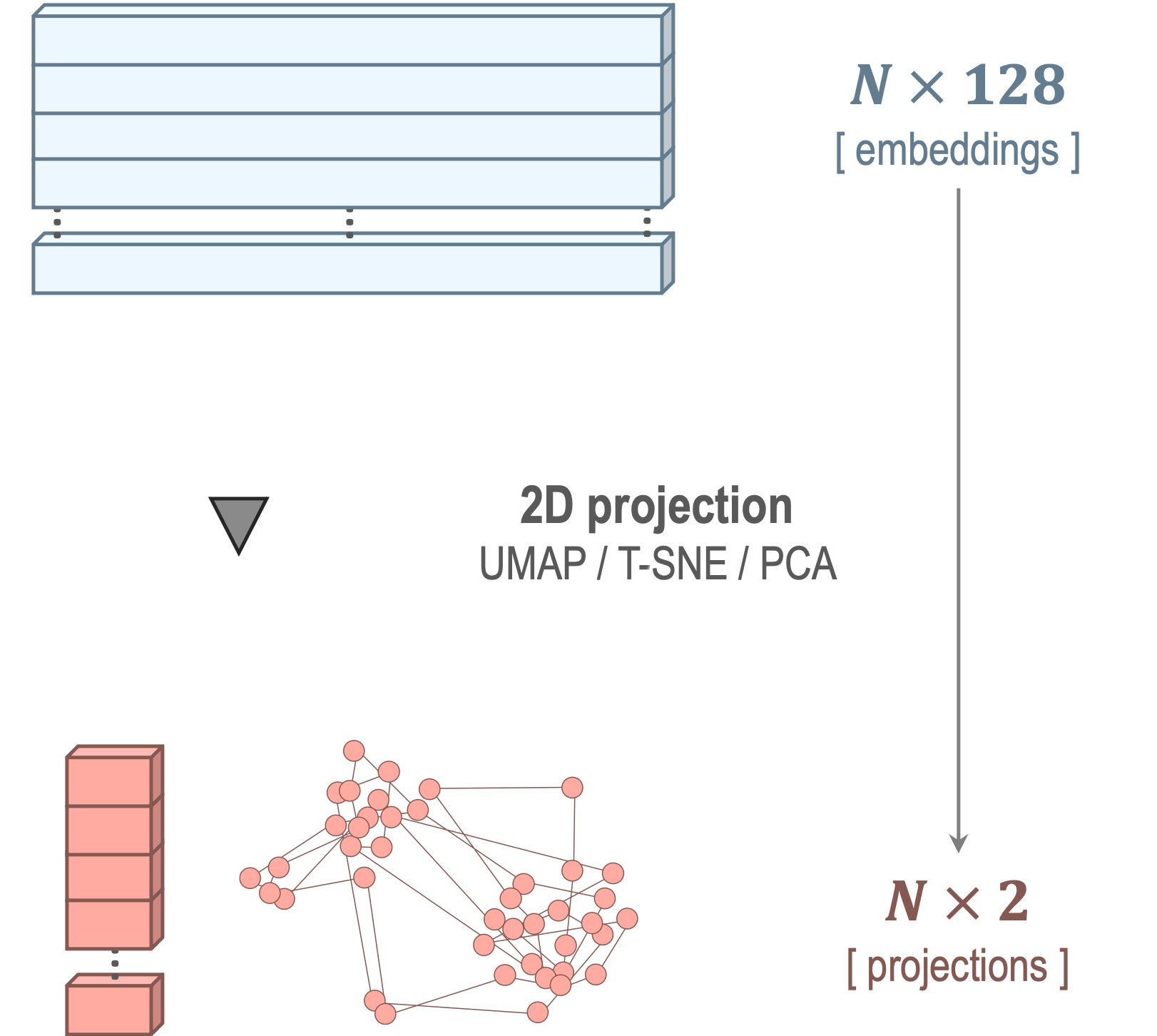}
\label{fig:compute_projections}
}
\caption{Inference process. (a) For each time window, the activation map of the last convolutional layer of the InceptionTime architecture is extracted and averaged according to the size of the sequence. (b) Taking as input all time windows, embeddings are extracted and projected, using UMAP, t-SNE or PCA.}
\end{figure}

With the configuration set in the control panel, the selected dataset is sliced into data windows through a sliding window process using the selected window size and the selected stride. Then, the whole set of windows produced is passed to the selected encoder. During this inference, we are not interested in the output of the model, but in its latent space, i.e., in its activations within one of its internal layers, which contain the encoded version (the embeddings) of each of the data windows. By default, if the proposed MTSAE is used as architecture for the encoder (See Section \ref{subsubsec:encoder}), the embeddings will be taken from the last layer of the encoder block, i.e., from the output of the last inception module from the InceptionTime architecture (See Figure \ref{fig:compute_embeddings}). Otherwise, in case a custom encoder is used, the layer from which to get the embeddings must be set. Normally, it will be the last layer before the head of the network, due to latter layers in a neural network having higher abstractions than earlier ones \cite{howard2020deep}.

% Shapes of the tensors through the process
Given a time series with $v$ variables, sliced into $N$ data windows of size $w$, the input tensor that will be passed to the model will have a shape of $N \times v \times w$. In the proposed InceptionTime-based MTSAE, the dimensionality of the embedding space at the last layer of the last inception module is 128 (that is, the number of filters of that layer is 128). There are no pooling layers in those modules that shrink the sequence length of the internal feature maps across the network. Therefore, at the end of the encoder section of the MTSAE the shape of the embeddings is $N \times 128 \times w$ (See Figure \ref{fig:compute_embeddings}). We average that tensor on the sequence dimension reducing it to $N \times 128$. At that moment, each data window is represented by a single embedding of 128 features. Finally, the selected dimensionality reduction algorithm is applied, projecting the embeddings into a $N \times 2$ table, ready to be visualised on a 2D canvas.

% Central plot and interactions it provides
The central plot to visualise projected embeddings is a connected scatter plot, as it was in the TimeCluster framework \cite{ali_timecluster_2019}. In this plot, data points represent the projection of the embedding of each data window, and straight lines are used to connect subsequent data windows to each other, in order to keep an idea of time within the plot. The size and colour of all the points is the same. The user can interact with this plot by zooming, cropping, select points and select areas of the 2D space.

\subsubsection{Time series plot}
Under the embedding projections canvas, we plot the time series dataset, and provide two-way interactions in both plots to explore and understand where the different parts of the embedding space are found on the original input space. When a point or area of the embeddings plot is highlighted, the time windows represented by those points are highlighted in the time series plot too, which allows the user to explain the patterns and anomalies of the embedding space based on the data (See the green area highlighted in Figure \ref{fig:gui}).

Other possible interactions available in the time series plot include zooming, cropping, and, in case of having multivariate time series, selecting a subset of the variables to be visualised, in order to reduce information overload in the plot.

\section{Experimental results} \label{sec:experiments}
% @dmontalvo:
% - Make experiments with the solar flux data (see description below), it would be nice to add the space projects to the acknowledgments
% The data is here (https://github.com/stardust-r/deep-learning-space-weather-forecasting/tree/master/data)
% - Make experiment to proof the usage of a multivariate time series? (Something in stumpy?, something synthetic?)
%TODO :Mention that UMAP is used with the default hyperparameters
In this section, we show the experimental results with the developed tool, with the aim of presenting the different applications of DeepVATS and the different scenarios to which it can be subjected. We start with the different datasets used in the experiments, and then we detail different experiments carried out to analyze the capabilities and limitations of DeepVATS.

\subsection{Datasets}
To put the tool into practice, we need time series datasets with different properties and from different domains. For this purpose, we used first synthetic time series, generated specifically for the direct purposes of the tool, to then move on to their real time series from different domains.

\subsubsection{Synthetic data}
\label{subsection:synthetic_data}

We constructed the synthetic time series as a linear combination of sinusoidal functions with different amplitudes and seasonalities. To this end, we assumed a discrete generation, taking minutes as the regular time step. We considered sinusoidal components for the hourly, daily, weekly seasonalities and different components for groups of $n$ hours, with $n$ taking the values of $2$, $3$, $4$, $6$, $8$ and $12$ hours. Based on these considerations, the generic function used to generate the synthetic time series is:

\vspace{-2mm}
\begin{equation}
    f\big(\,t\, | \, (\lambda_k)_{k\in\mathcal{M}},\, (\varphi_k)_{k\in\mathcal{M}},\, \gamma,\, \sigma\big) \, =\, \Bigg[\sum_{k\in\mathcal{M}} \lambda_k\cdot sin\bigg( \frac{2\pi}{60\cdot k}\, t + \varphi_k \bigg)\Bigg] + \gamma + \epsilon_t
    \label{eq:gen_function}
\end{equation}

\begin{figure}[h!]
    \centering
    \includegraphics[width=\textwidth]{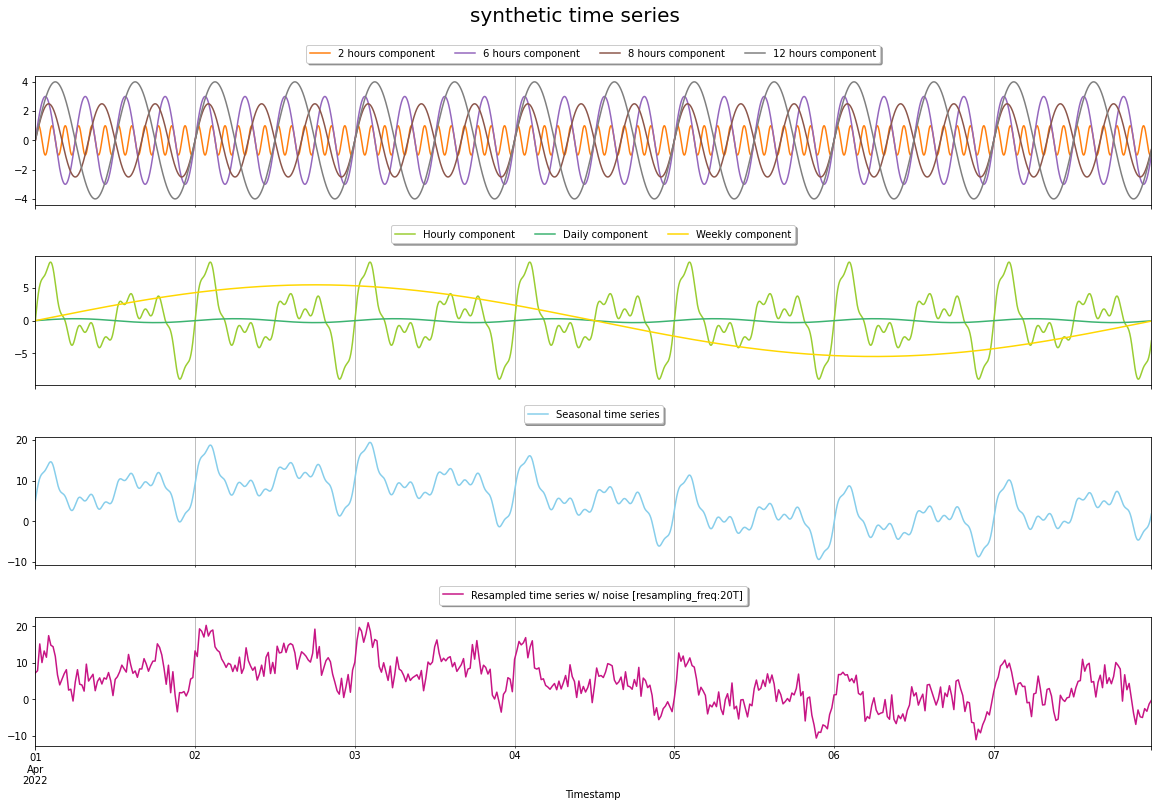}
    \caption{Example of a synthetic time series as a sum of sinusoidal components with different seasonalities and amplitudes. In this case, the generation parameters have been $\lambda_{2}=1$, $\lambda_{6}=3$, $\lambda_{8}=2.5$, $\lambda_{12}=4$, $\lambda_{24}=0.3$, $\lambda_{168}=5.5$, $\gamma=5$ and $\sigma=10$, without initial delay in the components and a resampled frequency of 20 minutes.}
    \label{fig:synthetic_data_example}
\end{figure}

\vspace{3mm}
\noindent where $\mathcal{M}=\{ 1,2,3,4,6,8,12,24 \}$ are the seasonalities (in hours) of the components, $\lambda_k$ the amplitudes of each component, $\varphi_k$ the initial delay of each component, $\gamma$ the \textit{offset} or initial displacement in the y axis, and $\epsilon_t \sim \mathcal{N}(0,\sigma^2)$ a stochastic white noise process. Figure~\ref{fig:synthetic_data_example} shows an example of generating a univariate synthetic time series, together with the decomposition of the sinusoidal functions that compose it.

With this generation technique, it is possible to synthesise both univariate and multivariate time series by generating the different variables independently. Additionally, we can concatenate time series generated with different parameters, thus giving rise to series with different stages. Following these procedures, we generated four synthetic time series datasets, whose equations and plots can be found in \ref{appendix:synthetic_datasets}:
\begin{itemize}
    \item \underline{S1}: a univariate time series spanning 24 days of data, made up of four segments of respective size $7.5$, $6.5$, $10$, and $4$ days, with different patterns constructed by varying the amplitudes of the different seasonal components. More specifically, two of them, the first and the third, maintain the same components with seasonality less than a day, and only the components with a daily and weekly seasonality vary, maintaining a very similar daily pattern in both segments. 
    
    \item \underline{S2}: a univariate time series with 20 days of data, built from a synthetic series used as a base, perturbing two segments of one hour each, at the beginning of the fifth day and at the end of the fourteenth, thus causing two anomalies in these segments of the time series.
    
    \item \underline{S3}: a univariate time series spanning 20 days of data, built with a single configuration of generation parameters, but with a new term in its definition to include a time-dependent increasing trend.
    
    \item \underline{S4}: a univariate time series spanning 20 days of data, constructed, like the series \textit{S2}, from a base synthetic series, but this time perturbing a single segment of one hour in the part end of the series, on the nineteenth day.
    
\end{itemize}

% Add multivariate dataset (from stumpy) (synthetic or real ¿?)
In order to test the capabilities of the tool to detect multidimensional time series motifs, we use a synthetic dataset called \emph{M-Toy}, gathered from the \emph{stumpy} library for time series data mining \cite{law2019stumpy}. It has three parallel variables, but only the first two are correlated and conform a multidimensional pattern. The third one is actually a ``random walk" that was purposely included as a red herring, not related to the other ones. This is meant to test whether the algorithm is robust to irrelevant dimensions and spurious data.

\subsubsection{Real-world data} \label{subsubsec:datasets-real}
We validated the capabilities of DeepVATS on real datasets for which we have access to expert information of what should be found on them. It should be noted that we selected them according to their purpose in the experimentation and not based on other factors such as their size, even though DeepVATS is especially useful for long time series. Below are brief descriptions of the three datasets employed, whose plots can be found in \ref{appendix:real}:

\begin{itemize}
    \item \underline{Arterial Blood Pressure (ABP)}: Firstly presented in \cite{gharghabi2017matrix}, it contains data from a healthy volunteer resting on a medical tilt table. At time 2400, the table was tilted upright, invoking a response from the homeostatic reflex mechanism, and producing two clear segments of data. %This dataset will be used to test the segmentation capabilities of DeepVATS.
    
    \item \underline{Parking}: Presented in \cite{Piccialli_Giampaolo_Prezioso_Crisci_Cuomo_2021}, it contains the hourly parking occupancy rate in 6 streets of Caserta and Naples, Italy. From them, we take the data of one of the streets, namely Piazza Vanvitelli. The parking occupancy rate is defined as the ratio between the number of occupied parking slots and the total number of parking slots present in a specific area. Hourly and daily seasonalities are present in the data.
    
    % \item \underline{F10.7}: Provided by The European Space Agency (ESA) Space Weather Service Network\footnote{\url{http://swe.ssa.esa.int/}}, this dataset contains the evolution of the solar flux at a wavelength of 10.7cm, an important space weather index to model and forecast in order to improve atmospheric drag calculations \cite{Stevenson_Rodriguez-Fernandez_Minisci_Camacho_2022}. This index, measured in solar flux units (sfu), has been stored continuously since 1947 by the Ottawa, and then Penticton Radio Observatories. Daily values of the F10.7 were taken to be those measured at 20:00 where appropriate, and missing values, more prevalent early in the time series, linearly interpolated.
    
    \item \underline{Kohl's}: The data contains a decade-long Google Trend query volume (collected weekly from 2004-2014) for the keyword Kohl’s, an American retail chain. The time series has a growing trend, and features a significant but unsurprising “end-of-year holiday bump”. We can see that the bump is generally increasing over time. This dataset was presented in \cite{matsubara2015web}, and used in the time series data mining library \texttt{stumpy} \cite{law2019stumpy} for the analysis of time series chains, i.e., motifs that evolve or drift in some direction over time.  

\end{itemize}

\subsection{Testing the capabilities for various time series data mining tasks}
In this section, we analyse the capabilities of DeepVATS to address three of most important time series data mining tasks: segmentation, recognition of repetitive patterns (or \textit{motifs}) and anomalies, and trend detection. The details of the hyperparameters used in the models trained  for the experiments of this section are listed in Table~\ref{table:params_data_mining}.

\begin{table}[h!]
\setlength\tabcolsep{2.4pt}
\centering
\bgroup
\small
\begin{tabular}{|c|c|c|c|c|c|c|c|c|c|} 
\hline
\multirow{2}{*}{\textbf{Task}} & \multirow{2}{*}{\textbf{Dataset}} & \multicolumn{3}{c|}{\textbf{Masking}} & \multicolumn{3}{c|}{\textbf{Window}} & \renewcommand{\arraystretch}{0.6}\multirow{2}{*}{\begin{tabular}[c]{@{}c@{}}\textbf{Batch}\\\textbf{size}\end{tabular}} & \multirow{2}{*}{\textbf{Epochs}} \\ 
\cline{3-8}
 &  & $\bm{r}$ & \textbf{\textit{stateful}} & \textbf{\textit{future}} & $\bm{w}$ & $\bm{w_{min}}$ & $\bm{w_{max}}$ &  &  \\ 
\hline
\multirow{2}{*}{Segmentation} & S1 & $0.4$ & Yes & No & 72 & 36 & 72 & 16 & 200 \\ % seg_synthetic
\cline{2-10}
 & ABP & $0.5$ & Yes & No & 220 & 200 & 220 & 32 & 50 \\ % seg_real_stateful
\hline
\renewcommand{\arraystretch}{0.6}\multirow{2}{*}{\begin{tabular}[c]{@{}c@{}}Patterns and \\anomalies\end{tabular}} & S2 & $0.5$ & No & No & 48 & 24 & 48 & 32 & 200 \\ % outliers_synthetic_not_stateful
\cline{2-10}
 & Parking & $0.5$ & No & No & 24 & 8 & 24 & 32 & 200 \\ % outliers_real_not_stateful_norm
\hline
\multirow{2}{*}{Trends} & S3 & $0.4$ & No & Yes & 96 & 32 & 96 & 32 & 200 \\ %trend_synthetic_future_low_w
\cline{2-10}
 & Kohl's & $0.4$ & No & Yes & 12 & 6 & 12 & 16 & 200 \\
\hline
Multivariate & M-toy & $0.7$ & Yes & No & 30 & - & - & 32 & 50 \\
\hline
\end{tabular}
\egroup
\caption{Training hyperparameters for the models trained to test the data mining capabilities of DeepVATS. In every case, the dimensionality algorithm employed was UMAP, with its default hyperparameters given in the official  implementation}
\label{table:params_data_mining}
\end{table}

\subsubsection{Segmentation} \label{subsubsec:seg}
Time series semantic segmentation, or just time series segmentation, is the task of dividing the series into homogeneous regions \cite{gharghabi2017matrix}. The timestamps when the time series changes from one segment to another are called \textit{change points}, and there are specific methods to detect them \cite{aminikhanghahi2017survey}. Here, we will show how DeepVATS allows extracting both the compact segments of the series and the change points.

\begin{figure}[h!]
    \centering
    \includegraphics[width=\textwidth]{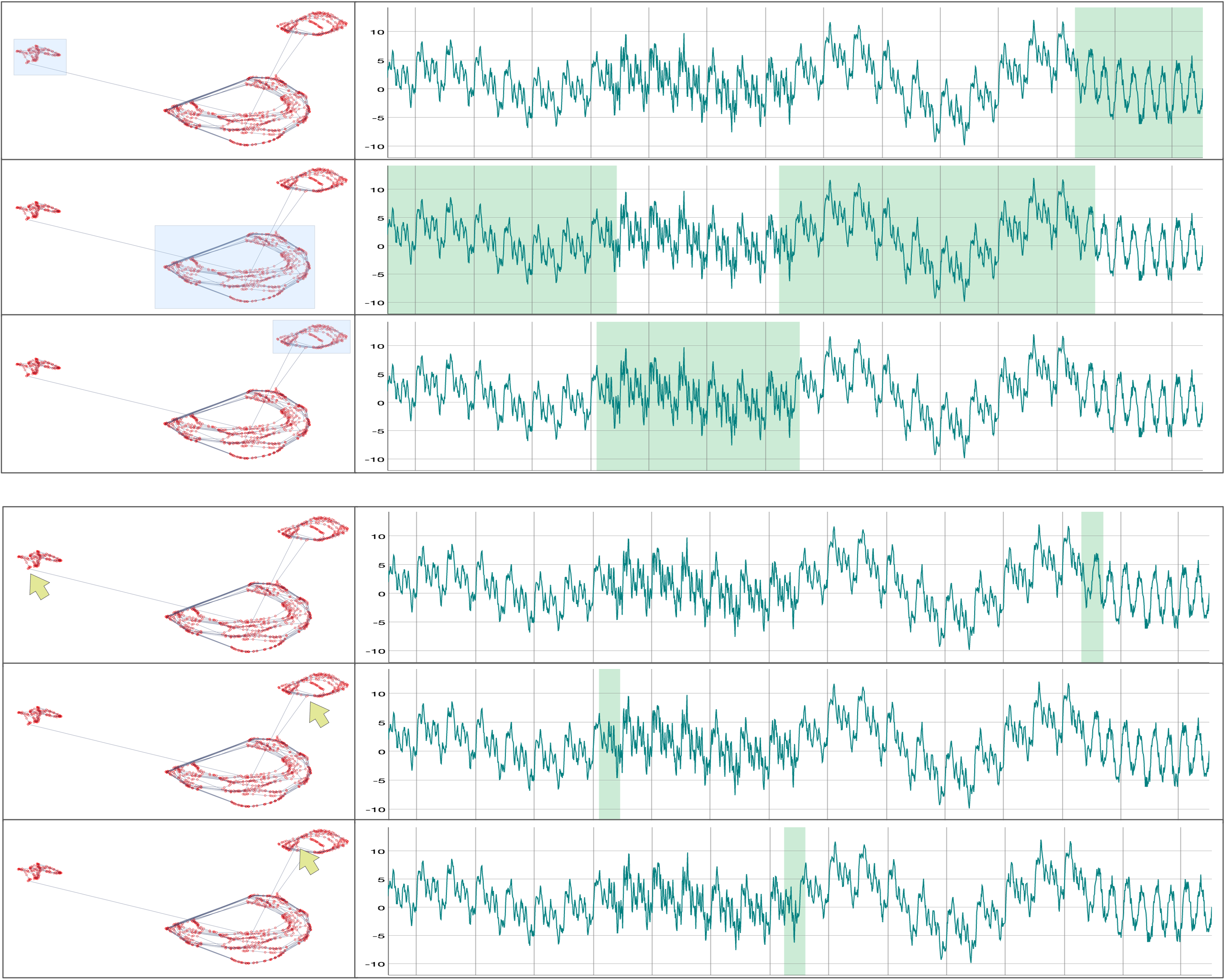}
    \caption{Segmentation and detection of change points of the synthetic series \textit{S1} with DeepVATS, using, in inference, a window size $w=54$, stride $s=2$ and UMAP as algorithm to project the \textit{embeddings}. On the left, the screenshots of the connected scatter plot with the projections of the \textit{embeddings}, in which the areas of points that have been selected to analyse their correspondence in the time series are marked in blue. On the right, the original time series with the areas corresponding to the points selected in the projections plot are marked in green. The first three representations show the detected segments, and the last three the change points among them, emphasised with yellow arrows.}
    \label{fig:seg_synthetic}
\end{figure}

Based on the first synthetic dataset (\textit{S1}), we trained a model in the deep learning module following a masking strategy with state (\textit{stateful}) and masking probability $r=0.4$, with variable window in the range $[36,72]$. Once trained, we loaded the model in the visual analytics module to explore its content. By exploring with the tool, we found that a window size $w$ of $54$ timestamps, a \textit{stride} $s$ of $2$ positions and UMAP as projection algorithm produced valuable insights. As it can be seen in Figure~\ref{fig:seg_synthetic}, the projection graph shows three clearly differentiated areas, with circular shapes, connected by gray segments. The circular shape of these clusters of points is no accident, and the reason for this is detailed in the next section. Each of these areas corresponds to a segment of the time series that have certain common characteristics. In the first three representations of the figure it can be seen that DeepVATS detected the three homogeneous segments that had been defined in the construction of the synthetic series, having considered the first and third fragments of the series as a single homogeneous segment, since both have common characteristics related to the seasonality of their components.

Since each point of the projection plot corresponds to a time window, and the points of contiguous windows are joined by a grey line, it is possible to detect \textit{windows with change points}, by looking at the points at the beginning and end of the line that join the different segments. The smaller the window size used in inference, the narrower the area of change between segments will be. In \ref{appendix:seg}, we complement this experiment with one analogous made on a real use case, namely the ABP dataset.

\subsubsection{Repetitive patterns and outliers}
In this section we aim at spotting repetitive patterns and anomalies with DeepVATS. On the one hand, anomaly (or outlier) detection refers to the task of finding points or subsequences in the time series that differ from the common behaviour of the rest of the data. On the other hand, the recognition of repetitive or cyclical patterns involves finding subsequences that appear recurrently throughout the time series, due to either the seasonal nature of the series or to the presence of a \textit{motif}, .i.e, a previously unknown pattern that carry precise information about the underlying source of the series \cite{mueen2014time}.

\begin{figure}[h!]
    \centering
    \includegraphics[width=\textwidth]{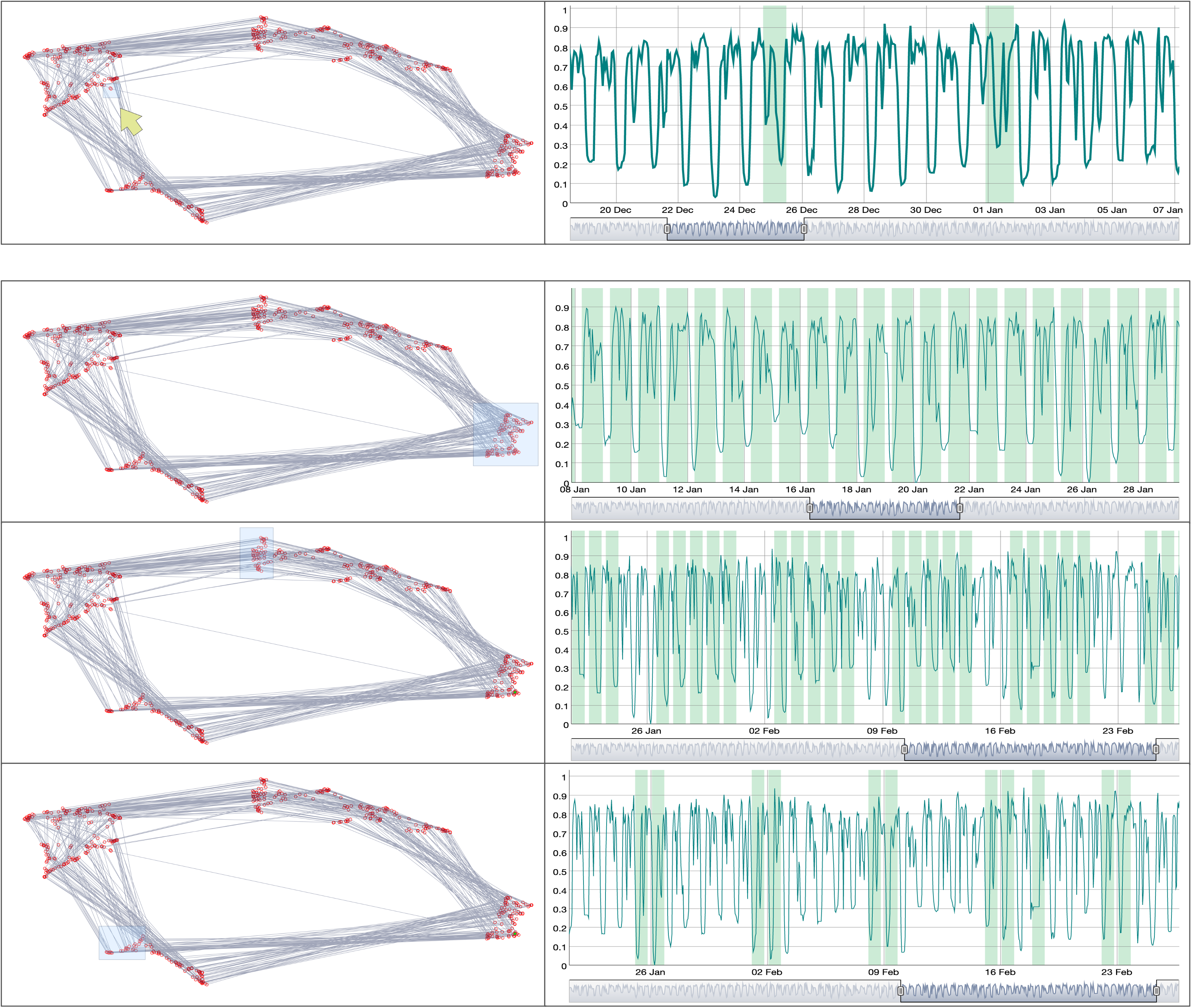}
    \caption{Anomaly detection (first screenshot) and repetitive patterns (last three screenshots) on the dataset of parking occupancy with DeepVATS, using, in the VA module, a window size $w=19$, stride $s =4$ and UMAP as dimensionality reduction technique. Model training parameters are summarised in Table \ref{table:params_data_mining}.}
    \label{fig:outliers_motifs_real}
\end{figure}

To account for this analysis, we trained MTSAEs on two datasets, namely \textit{S2}, built specifically for this purpose, and \textit{Parking} (See Section \ref{subsubsec:datasets-real}). Here we show the results on \textit{Parking}, leaving the results on the synthetic data in \ref{appendix:motifs}. We trained the encoder following a stateless masking strategy with $r=0.5$ and a variable window size of $[8, 24]$, given that the series have a time step of 1 hour and presents daily seasonalities. Then, we explored the model in the VA module and found the results shown in Figure~\ref{fig:outliers_motifs_real}. The first aspect to highlight in the projections plot is that points are distributed following an elliptical shape, which indicates that the model has been able to detect the cycles present in the data (this is in accordance with the results found in \cite{ali_timecluster_2019}). However, we can see several grey segments that do not follow the elliptical path and cross the middle of the figure. Examining the windows corresponding to the extreme points of these segments, we see that they correspond to time windows that contain the days December 25 and January 1 respectively. On those days, the parking occupancy followed a different pattern, and in fact, if we observe the series those days in an extended way (see the right graphic representation with the highlighted windows in green on the time series) we can confirm that its behaviour differs from the surrounding days. Outliers, therefore, appear in DeepVATS as points that do not follow the path common to the rest of the time windows.

Considering the groups of points formed along the cyclic trajectories, the second screenshot of Figure~\ref{fig:outliers_motifs_real} shows how all the points located in the lower right part of the trajectory correspond to the pattern of occupation of the daytime hours, leaving outside the night hours. The following screenshot of this same figure highlights certain points in the projection space that represent occupancy patterns on weekdays, while the last one shows the points that correspond to the repetitive pattern of weekends and some holidays. Therefore, we can conclude that clusters within the elliptical path correspond to windows that follow the same recurrent pattern, beyond the seasonality periods of the series.

% TODO: This would link very well with a comparative to DCAE
The masking strategy that has provided the best results for detecting anomalies and repetitive patterns is the default stateless masking strategy. In these cases, the probability of masking follows a uniform distribution, increasing the probability that isolated timestamps are masked. This helps the \textit{encoder} to identify point-based anomalies in and not ignore them when it comes to reconstructing the masked parts in the \textit{decoder}, as it happens with classic autoencoders.

\subsubsection{Trend detection}
Trend detection in a time series tries to find significant and prolonged changes over time. Some classical approaches accomplish this task using statistical tests \citep{onoz2003power}. In DeepVATS, we will try to detect these changes from the shapes that follow the trajectories of points in the projection space. Among the different masking options available to train models in DeepVATS, we found that the \textit{future} masking strategy (See Figure~\ref{fig:masking_strategies}) is the one that best suits this task. A possible explanation for this is that having to reconstruct the latest values of the time windows based on previous knowledge makes the model learn to interpret the trends in the data in order to make the reconstruction of the future values take those trends into account.

\begin{figure}[h!]
    \centering
    \includegraphics[width=\textwidth]{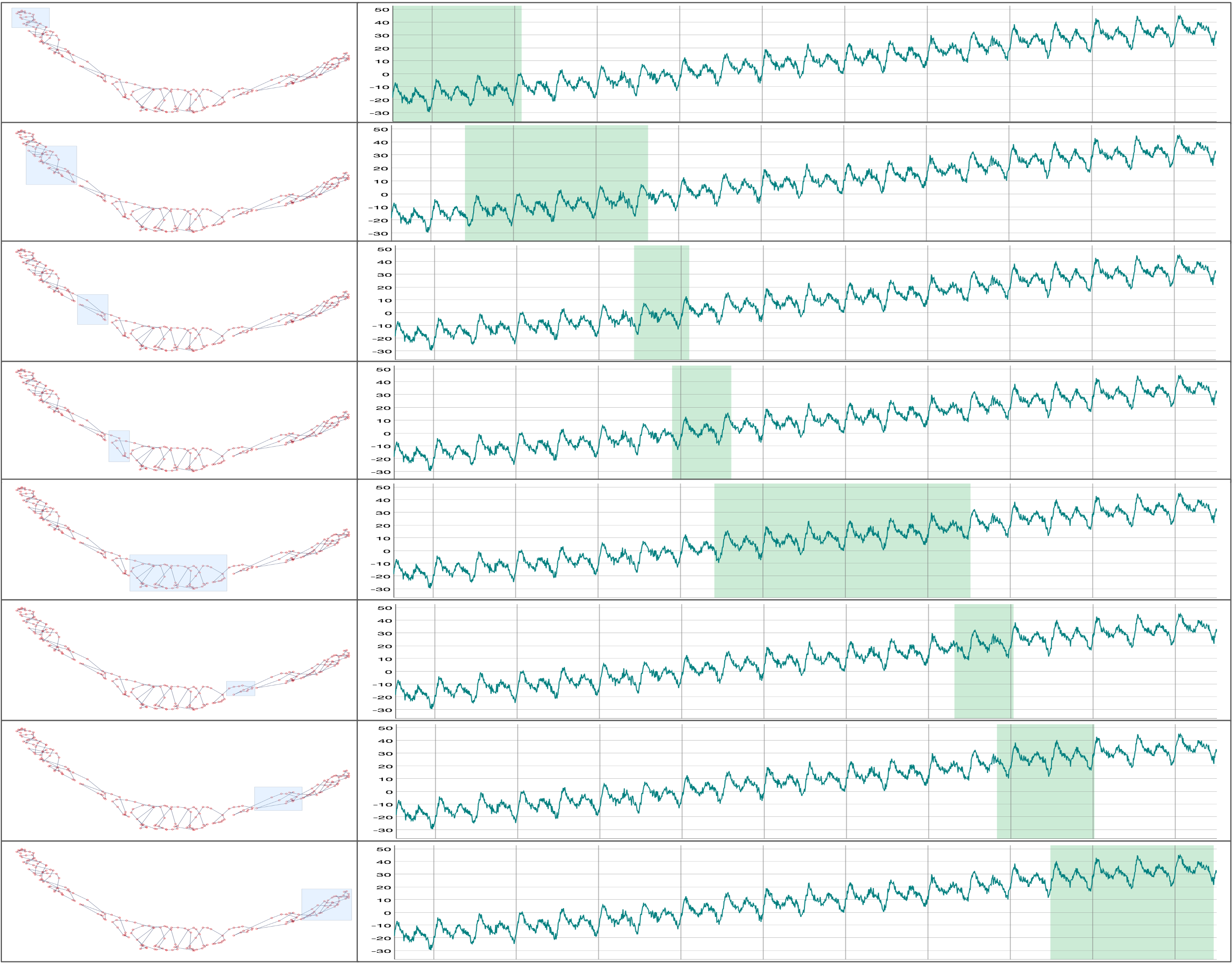}
    \caption{Trend detection on synthetic dataset \textit{S3} with DeepVATS, using, in the VA module, a window size $w=50$, stride $s=9$ and UMAP as dimensionality reduction technique.}
    \label{fig:trend_synthetic}
\end{figure}

As a first approximation analysing the ability of DeepVATS to detect trends in the data, we trained a model with the synthetic dataset \textit{S3}. We employed a \textit{future} masking strategy  and a variable window size in the range $[32,96]$. The results after exploring the model in the VA module are shown in Figure \ref{fig:trend_synthetic}. Although the series is cyclical, the trajectories in the projection space do not seem to have an elliptical or circular shape (as  happened with the models of the previous section), being the points distributed in a sort of parabolic trajectory. However, if we carefully examine the groups of points within this trajectory and the segments that join them, we appreciate how they retain a certain circular shape, with the centre of the circle of each cycle shifting as the series progresses (see carefully screenshots 3, 4 and 6 in Figure~\ref{fig:trend_synthetic}). The projections contain the cycles, but display them in displacement as a result of the trend of this series. Therefore, although a certain idea of trend is captured in the two-dimensional space, it is not reflected as clearly as it has occurred with anomalies, patterns and segments.

\begin{figure}[h!]
    \centering
    \includegraphics[width=\textwidth]{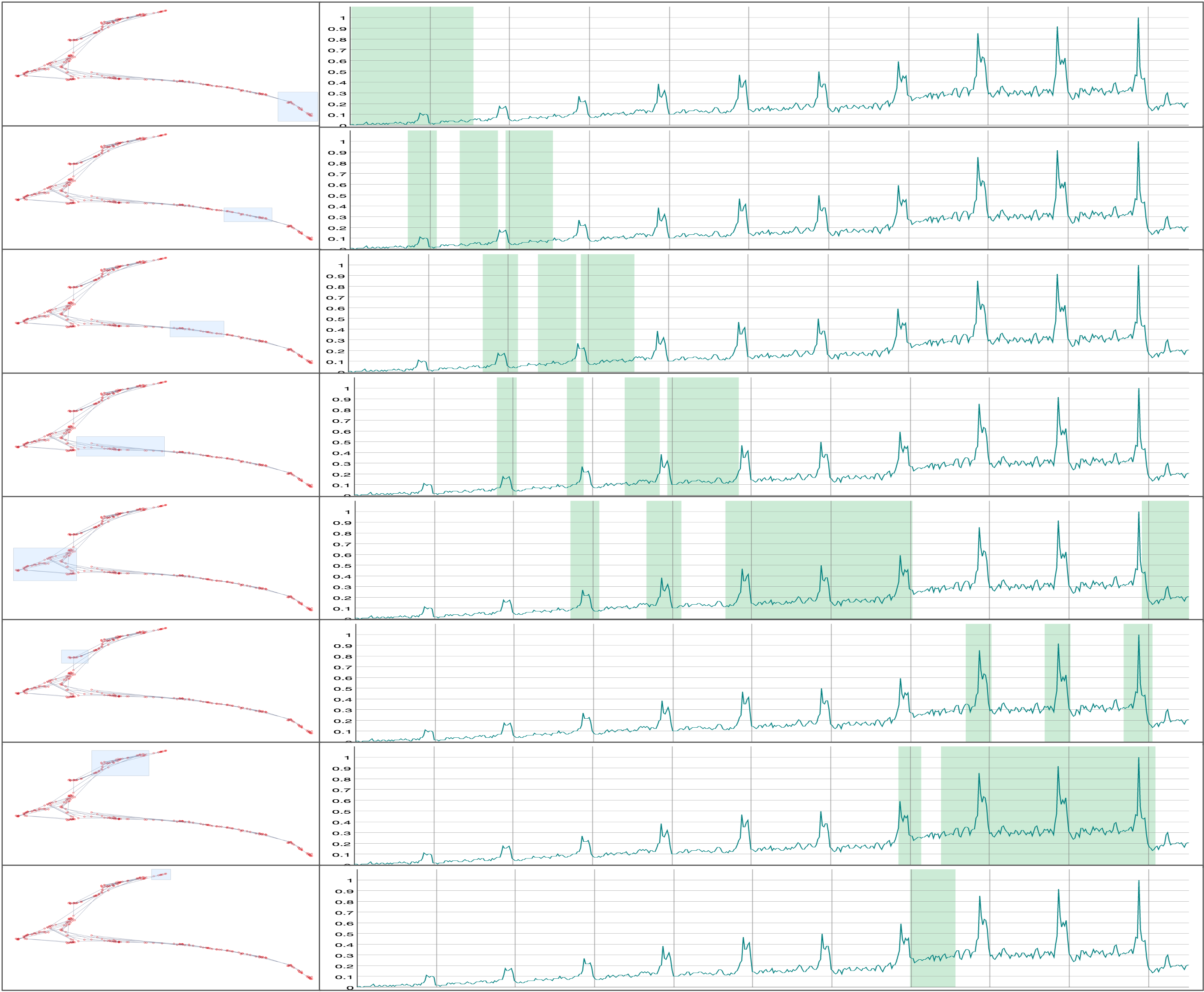}
    \caption{Trend detection on the \textit{Kohl} series with DeepVATS, using, in the VA module, a window size $w=12$, stride $s=2$ and UMAP as dimensionality reduction technique.}
    \label{fig:trend_real}
\end{figure}

We corroborated this hypothesis after applying the tool to the \textit{Kohl} dataset (See Section \ref{subsubsec:datasets-real}). We trained the MTSAE on this dataset employing a future masking strategy, analogous to that applied in the synthetic use case above, and a variable window size of $[6,12]$ timestamps. The results after exploring the model embeddings in the VA module are shown in Figure~\ref{fig:trend_real}. Unlike S3, \textit{Kohl} is a time series whose amplitude on each cycle is not constant. In addition to having a growing global trend, it grows in amplitude each year. This fact is translated into a projection space in DeepVATS that provides little valuable knowledge about the series. As it can be seen in Figure~\ref{fig:trend_real}, the projected points seem to follow a linear trajectory, but if we look at the time windows corresponding to some of the groups of points that are close in the trajectory, we don't see a clear mapping between them and continuous segments of the original series. Furthermore, the points associated to time windows containing peaks are not clustered together in a region of the projection space. In conclusion, we cannot extract valuable insights about trends using DeepVATS, as it is developed so far.

\subsubsection{Multidimensional patterns}
Finally, we test the capabilities of DeepVATS to analyse multivariate time series. Note that we do not have to change anything from the pipeline used for the univariate analysis above, since both the neural architecture and the masking strategy used to create the backbone MTSAE work for multivariate time series out of the box. By default, DeepVATS creates unsynchronised masks to train the MTSAE, which means that the same time step can be part of the mask or not depending on the variable we are focusing on. This naturally makes the training robust to spurious dimensions and allows the network to find patterns between a subset of dimensions. However, this can be changed as part of the configuration, making the mask synchronised across all variables, which is only recommended in case one knows in advance that all dimensions of the series are involved in the patterns of interest.

\begin{figure}[h!]
    \centering
    \includegraphics[width=\textwidth]{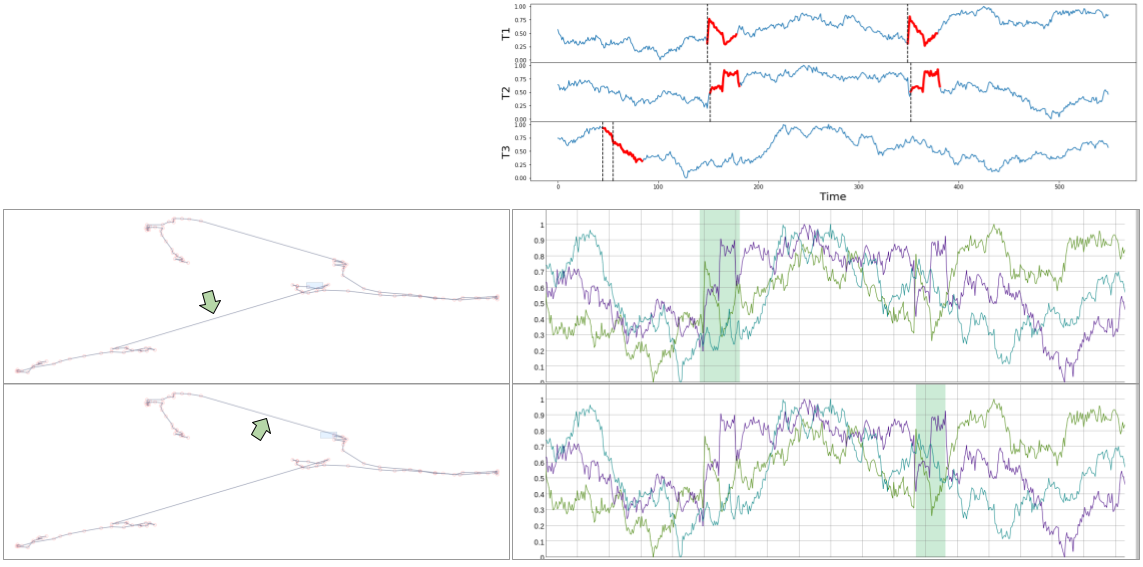}
    \caption{Multivariate analysis on the \textit{M-toy} series with DeepVATS, using, in the VA module, a window size $w=30$, stride $s=5$ and UMAP as dimensionality reduction technique. The plot of the first row has been gathered from the \emph{stumpy} documentation to see the multidimensional pattern that we are looking for (the one in T1 and T2). The colour map in the time series plots of the second and third rows are: (T1: green, T2: purple, T3: blue)}
    \label{fig:multivariate}
\end{figure}

We trained a MTSAE on the \emph{M-toy} dataset (see Section \ref{subsection:synthetic_data}, using a fixed window size of $w = 30$, as it is done in the \emph{stumpy} library from which the dataset was gathered. The results after exploring the model embeddings in the VA module are shown in Figure~\ref{appendix:motifs}, along with a screenshot from the \emph{stumpy} documentation \footnote{\url{https://stumpy.readthedocs.io/en/latest/Tutorial\_Multidimensional\_Motif\_Discovery.html}} with the results of running a 1-dimensional matrix profile based motif discovery algorithm on the dataset. This algorithm resulted in a real multidimensional motif found between the first two channels of the series (T1 and T2), and a spurious motif in channel T3, which we would like to avoid because, as it was said in Section \ref{fig:synthetic_datasets}, that channel was artificially introduced as noise.

As it can be seen in the projections of Figure \ref{fig:multivariate}, there are two long segments in the connected scattered plot with no point in between, which represent ``irregularities" in the embedding space. In fact, by analysing the points at the ends of each segment, we can see that they correspond to the two appearances of the motif that we were looking for (the one between T1 and T2). Since there are no other irregularities in the embedding space, we can conclude that DeepVATS has helped to easily spot this motif, and it has avoided the effects of noisy variables such as T3.

\subsection{Influence of using variable window size during training}
\label{subsection:training_variable_window_size}

The window size is a critical hyperparameter in the construction of models that work with long time series, regardless of the task they perform. The ideal window size depends on the dataset itself, and it is common to have it set manually by an expert \cite{gharghabi2017matrix}. In this work, the use of a neural architecture in which the number of trainable parameters does not depend on the window size (see Section \ref{subsubsec:encoder}) makes it possible to train the models with different window sizes. This approach, in theory, should reduce the impact of choosing the optimal window size in order to obtain the expected results.

\begin{figure}[h!]
    \centering
    \includegraphics[width=\textwidth]{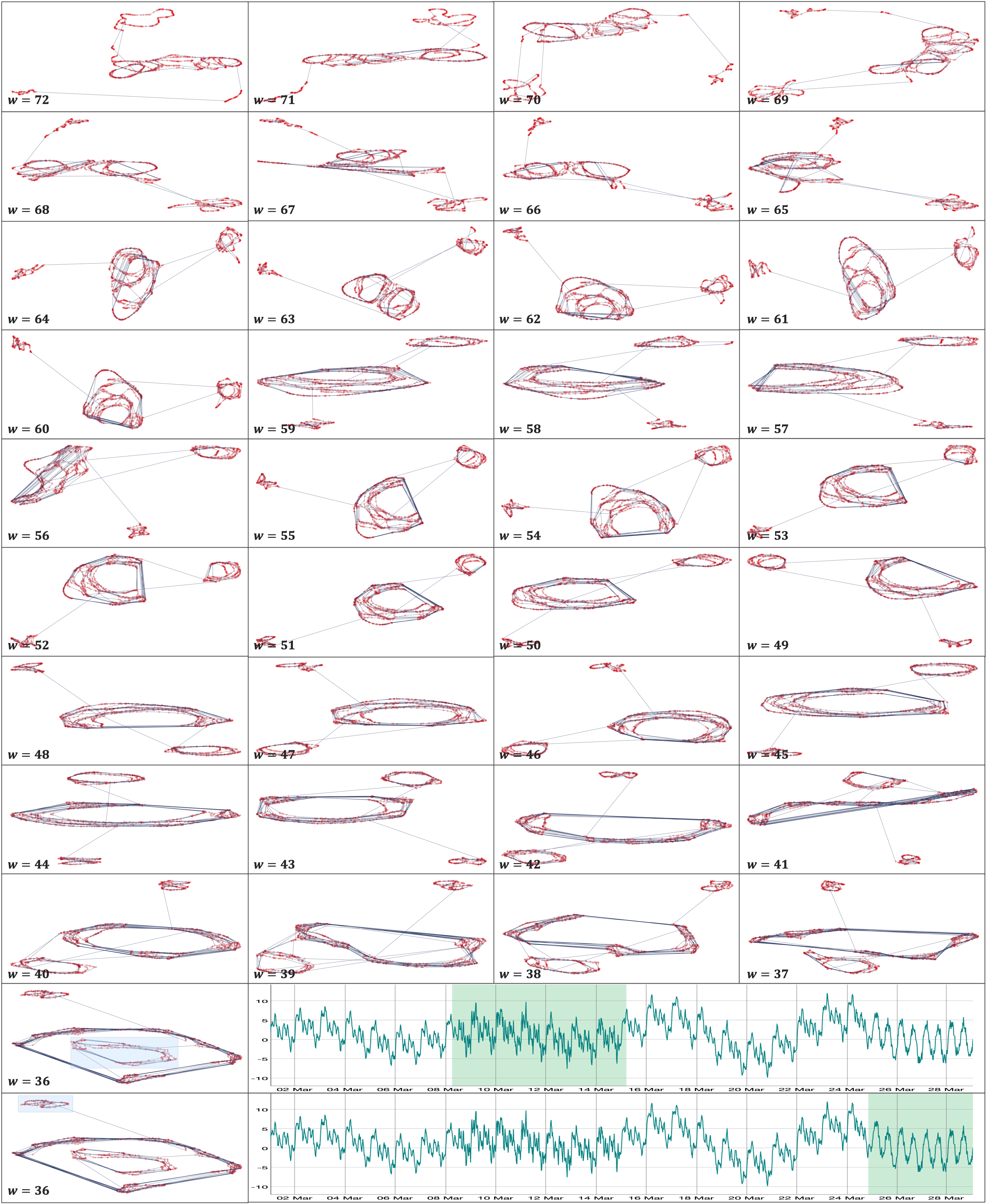}
    \caption{Influence of the window size selected in the VA model for a model trained with a variable window size of $[36, 72]$. To obtain the plots, we fixed a \textit{stride} $s$ of $2$ positions and UMAP as reduction technique. The segmentation applied to the case of size $w=36$ is shown at the bottom.}
    \label{fig:seg_synthetic_variable_w}
\end{figure}

To demonstrate this hypothesis, we compared the results, for a segmentation task, between two MTSAE models trained on the same data set (S1), one trained with a variable window size in the range $[36,72]$ (the same model as in \ref{subsubsec:seg}) and the other with a fixed window size. Figure~\ref{fig:seg_synthetic_variable_w} shows the results of exploring the model trained using variable window size. We took screenshots of the projection plot obtained for every value of $w$ in the range $[36, 72]$ selected in the VA module, simulating the behaviour of a user that wants to analyse a dataset without a prior knowledge of the ideal window size. As it can be observed, in all the plots the three blocks of points corresponding to each one of the homogeneous segments of the series can be clearly appreciated.

\begin{figure}[h!]
    \centering
    \includegraphics[width=\textwidth]{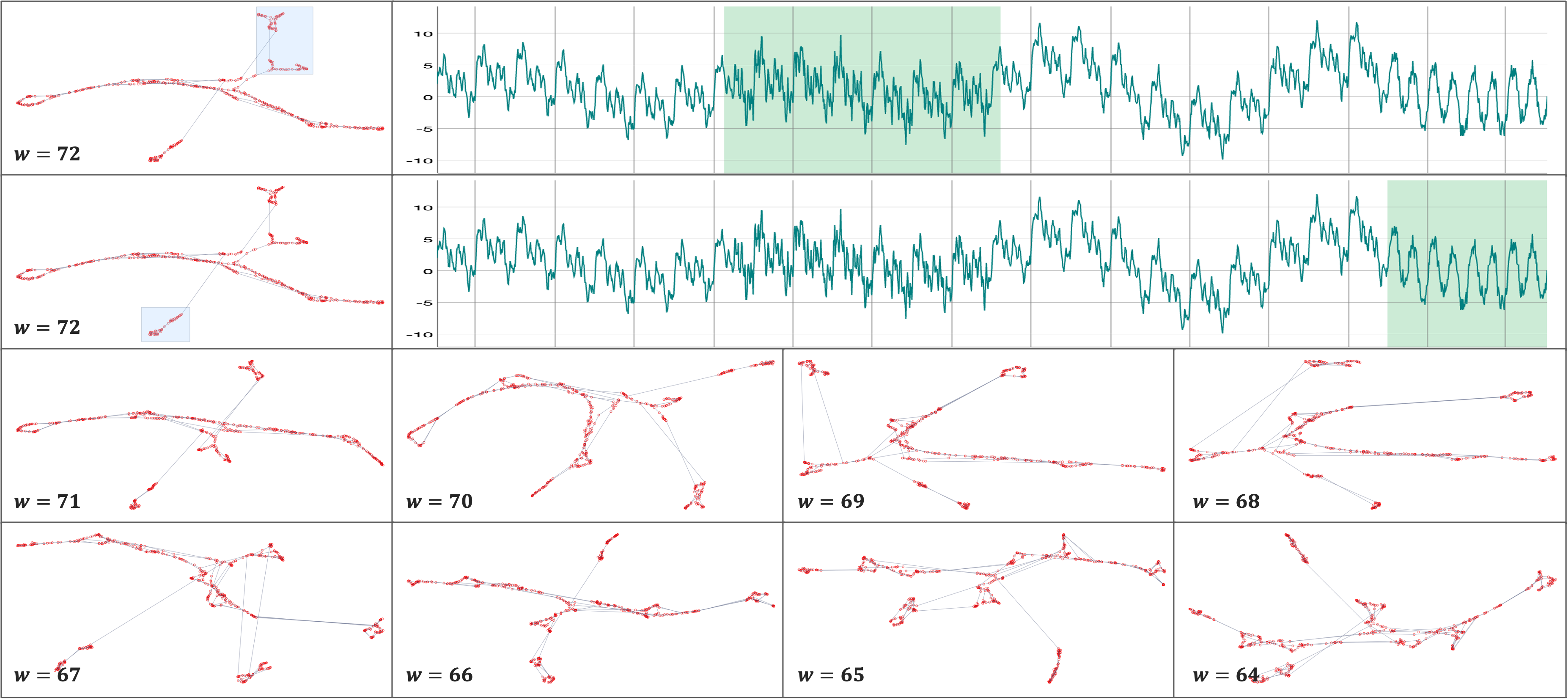}
    \caption{Influence of the window size selected in the VA module with a trained model with a fixed window of $72$ timestamps. To obtain the plots, we fixed a \textit{stride} $s$ of $4$ and UMAP as projection algorithm. At the top is the segmentation applied to the case $w=72$.}
    \label{fig:seg_synthetic_fixed_72}
\end{figure}

Regarding the fixed window size case, we trained three models with $w=32$, $w=54$ and $w=72$. We used the same training configuration in all of them as the previous segmentation model except for the window size. From these three models, the one that provided the most valuable insights in the VA module to segment the series was the one trained with a window size of $72$. In Figure~\ref{fig:seg_synthetic_fixed_72} we can see the scatter plots obtained with this model in the VA module, varying the selected window size from $w=64$ to $w=72$. As it can be seen, the clusters of points associated to the series segments are generally less obvious than in the previous case with the variable window size, and in fact, with $w=64$ the plot is confusing. In addition, these models do not show the cyclic behaviour of the series in the form of elliptical or circular paths, as it happens with the model trained with a variable window size.

\subsection{Generalisation capabilities in the online mode}

\begin{figure}[h!]
    \centering
    \includegraphics[width=\textwidth]{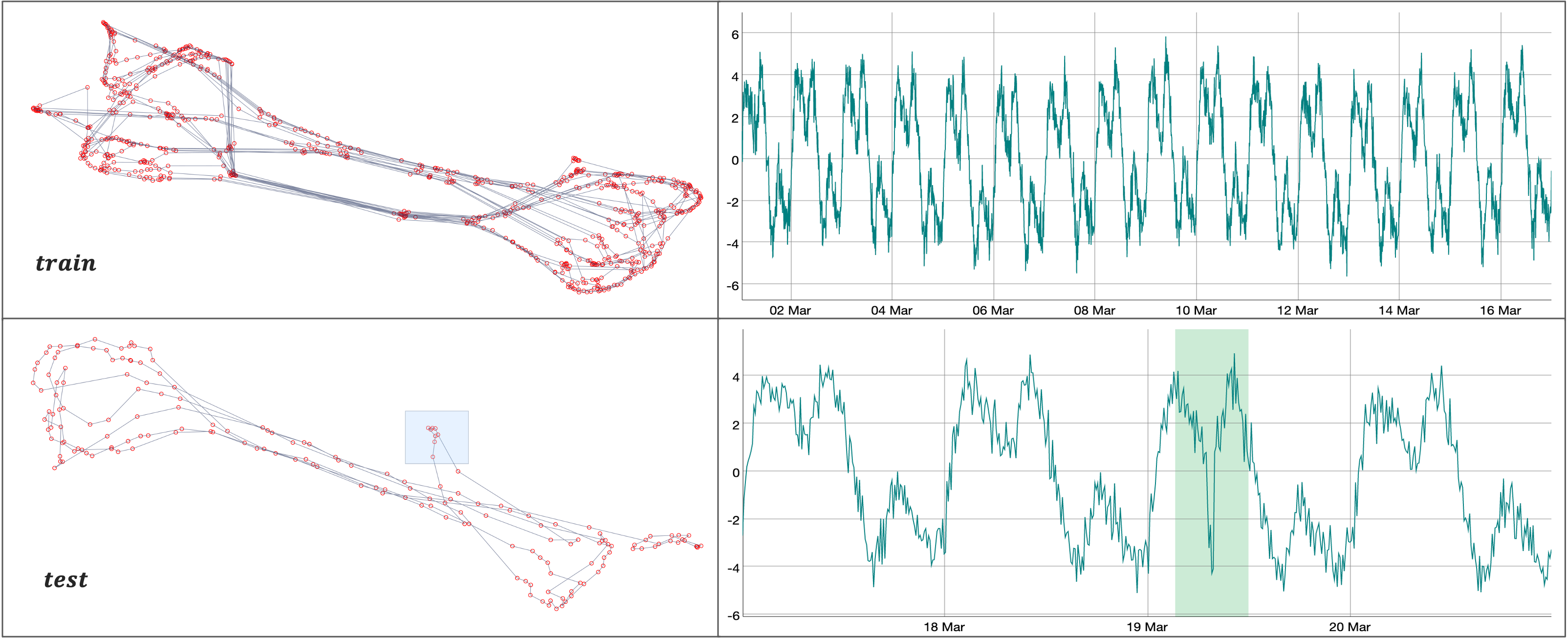}
    \caption{Anomaly detection in the \textit{online} analysis mode. The upper screenshot shows the projected embeddings of the training time series. The lower one shows the analogous plot of the series reserved for testing, marking the area of points corresponding to the anomaly. In both screenshots, we configured the VA module with $w^{\textnormal{VA}}=33$, $s=3$ and UMAP  as dimensionality reduction algorithm.}
    \label{fig:online_synthetic}
\end{figure}

In this section, we analyse whether the models trained in DeepVATS can maintain their performance when the MTSAE is not trained on the whole series but only in a subset of it, to which we referred to as ``online analysis" in the DL module, given its applicability to online analysis where new data from the same process arrives continuously, or when the input series is too long to be used entirely to train the model. To do this, we used the synthetic dataset S4, which has a disturbance in the final part of the series. We split it into two sets, one training set with the initial 80\% of the series and one test set with the remaining 20\%, that contains the disturbance. We logged both datasets in the storage module and trained the encoder model in the DL module using just the training set, and following the ``online analysis" mode shown in Figure~\ref{fig:offline_online}. We employed a stateless masking strategy, which was found to be helpful to spot outliers in the previous section, and a variable window size of $[30, 60]$. Figure~\ref{fig:online_synthetic} shows the projected embeddings for both the training and test datasets using the same model. In the test embeddings, we observe how a group of points deviates from the main trajectory, and analysing where in the time series those points come from, we see that their corresponding time windows contain the disturbance. This demonstrates that the knowledge contained in the trained model is generalisable to new streams of data, allowing a fast analysis of new series without the need of retraining the model.

\subsection{Comparison to related work}
Finally, we conclude the experimentation with a comparison of the latent spaces produced by DeepVATS with respect the following methods (already introduced in the related work):
\begin{itemize}
    \item \emph{TimeCluster} \cite{ali_timecluster_2019}, which shares the same goal of DeepVATS, but is internally built on top of a different neural network architecture. More specifically, TimeCluster's latent space is based on a Deep Convolutional AutoEncoder (DCAE), trained without any masking strategy.
    
    \item Multi-Dimensional Scaling (MDS) \cite{torgerson1952multidimensional}, on top of which Bach et al. developed their \emph{TimeCurves} \cite{bach2015time} \footnote{We do not apply the full TimeCurves method due to it is meant for irregular time series, whereas DeepVATS is meant for regular ones}. MDS is a statistical technique used to visualise the similarity or dissimilarity between objects in a multi-dimensional space by representing them in a lower-dimensional space (usually 2D or 3D) while preserving their pairwise distances as much as possible.
\end{itemize}

% Comparison figure
\begin{figure}[ht]
  \centering
  \subfigure[Latent space of TimeCluster]{
    \includegraphics[width=\textwidth]{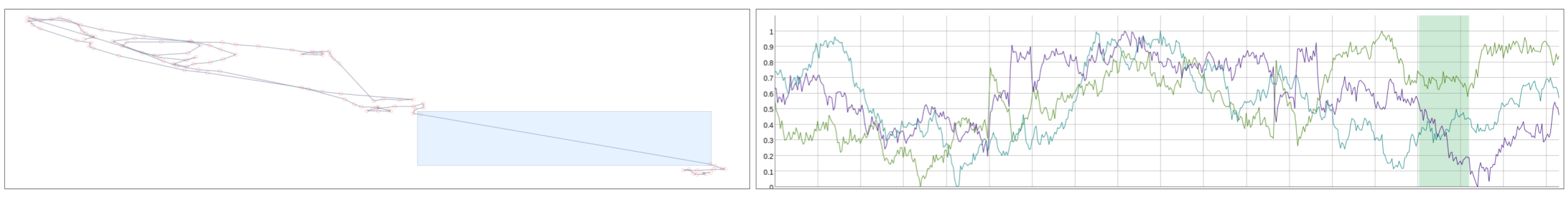}
    \label{fig:subfig1}
  }
  \subfigure[Latent space of MDS]{
    \includegraphics[width=\textwidth]{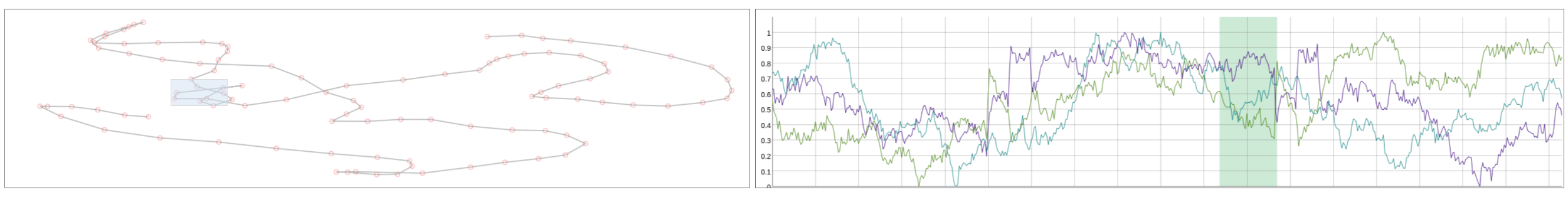}
    \label{fig:subfig2}
  }
  \caption{Latent spaces of both DCAE and MDS on the M-toy dataset, using a window size of $w=30$ and a stride of $s=5$, to compare the results with the ones shown in Figure \ref{fig:multivariate}. On the right of each latent space, the M-toy dataset is plotted, along with a green rectangle showing the highlighted section of the latent space.}
  \label{fig:comparison}
\end{figure}

% Dataset M-toy used for the experimentation
The dataset used for the comparison is \emph{M-toy}, which is arguably the most challenging dataset used in this experimentation, due to it is multivariate and contains spurious variables that challenge the robustness of the algorithms. In Figure \ref{fig:multivariate}, we showed that DeepVATS' latent space could successfully spot the two appearances of the motif formed by variables T1 and T2 of the dataset, ignoring the spurious variable T3. That Figure can be compared side by side with Figure \ref{fig:comparison}, in which the analogous latent spaces for the dataset under the same pre-processing conditions (that is, $w=30$, $s=5$ and UMAP for dimensionality reduction in the case of TimeCluster) are displayed for both the TimeCluster and MDS methods. For TimeCluster, the neural network autoencoder has been trained using the exact same architecture used in the paper, with the same number and configuration of layers (See Table 1 in \cite{ali_timecluster_2019}). Note that, given the modularity of DeepVATS, we have been able to add this architecture to the tool and explore its latent space interactively without changing anything in the pipeline, and in that way we could add and try any other autoencoder. In the case of MDS, due to it is not a deep learning algorithm, we have run it in a separate pipeline (included in the repository as well) using \textit{scikit-learn}'s implementation. A multivariate Euclidean distance has been used to create the dissimilarity matrix that is fed into the algorithm.

% results
As it can be seen, none of the latent spaces shown in Figure \ref{fig:comparison} show clearly any appearance of the motif formed by the variables T1 and T2 (See the motif in Figure \ref{fig:multivariate}). In the case of TimeCluster, the latent space has one clear irregularity (See the blue rectangle in \ref{fig:subfig1}, but when analysing it in the VA module, we see that it maps to a different part of the time series, far from the motif of interest. On the other hand, in the latent space of MDS there's not even a clear irregularity found among the points of the embedding space. There is, however, a small concentration of points (See the blue rectangle in \ref{fig:subfig2}) that stands out of the rest of the curve, but again, when mapping it to the time series we can see that it does not correspond at all to the motif of interest. We can then conclude that, for this challenging use case, DeepVATS' MTSAE is the only method whose embeddings are rich enough to spot the motif hidden in the dataset. We argue that the reason behind this is the inclusion of masking strategies in the neural network training, as well as the flexibility provided by using with variable window sizes during training. We leave as future work the development of more comparison studies, which we will include in the repository of the tool.

\section{Conclusions and Future Work} \label{sec:conclusions}
% FW: Datasets of multiple time series
% FW: AutoML for training the encoder
% FW: 3D projections
% FW: XAI to select automatically variables. This helps especially with high-dimensional time series
% FW: Add time encodings and matrix profile as additional variables to the model
% FW: Explore trend detection further
In this paper, we present the tool DeepVATS, which trains, in a self-supervised way, a Masked Time Series AutoEncoder (MTSAE) that reconstruct time windows of a long time series, and projects the knowledge contained in the model's latent space (i.e., the embeddings) into an interactive graphical user interface, which allows to extract, in a fast way, high level information about the structure of the series, such as homogeneous segments, repetitive patterns, and outliers. The tool is publicly available \ref{fn:dvats-github} and has been built with a modular design that encourages its extension.

\begin{figure}[h!]
    \centering
    \includegraphics[width=\textwidth]{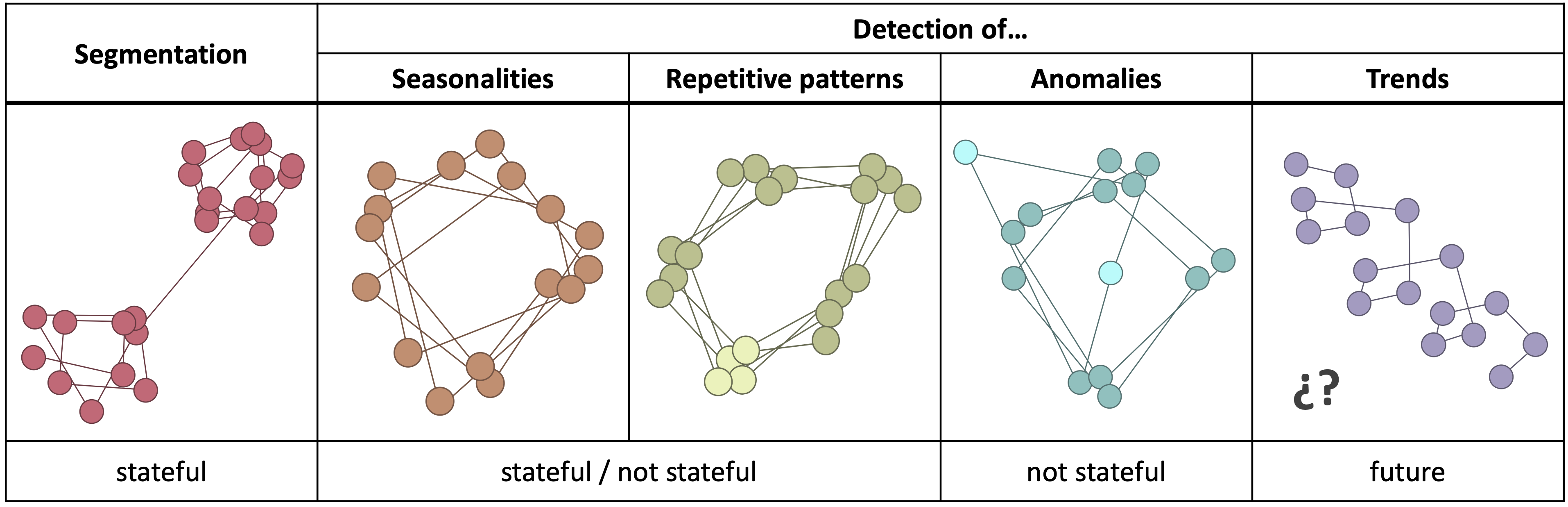}
    \caption{Summary of DeepVATS capabilities for time series tasks, along with the visual pattern associated to them in the VA module, and the most suitable masking strategy to be used in the DL module.}
    \label{fig:conclusions}
\end{figure}

In view of the experiments carried out, we conclude that DeepVATS has great potential to provide valuable insights in long cyclical time series for tasks such as semantic segmentation, detection of seasonalities, recognition of repetitive patterns and anomalies (See Figure \ref{fig:conclusions}). Using a single connected scatter to display the projected embeddings of the model as a two dimensional trajectory, segments can be visually spotted as clusters of independent points, seasonalities come from the shape of those clusters, repetitive patterns as compact groups within the trajectory, and anomalies as points that do not follow the nominal trajectory. For other tasks, such as trend detection, we did not found enough evidence of a visual pattern within the tool.

%TODO: This only makes sense if we show a comparison in the experiment section
With regard to masking strategies, we showed that having flexibility in the masking configuration allows DeepVATS to perform well on a greater range of tasks. Compared to the \textit{TimeCluster} \cite{ali_timecluster_2019} tool, the use of a stateless masking strategy that forces the model to reconstruct isolated timestamps instead of complete time windows, allows to address the task of anomaly detection with a higher degree of accuracy. On the other hand, the stateful masking strategy helped the models to detect homogeneous segments within the series, while the strategy based on masking future values helped to extract some indications of trend in the non-stationary series.

On the other hand, based on the results obtained in Section \ref{subsection:training_variable_window_size}, we showed that the approach for variable window size training presented in this project provides the tool with a lower dependency on the choice of the optimal window size. While in other approaches, such as \cite{gharghabi2017matrix}, this value must be provided by domain experts for the results to be acceptable, DeepVATS can train a robust and effective model can be trained just setting a range of potential window sizes, without the need to know the exact optimal value.

Another remarkable aspect of DeepVATS lies in its modular implementation, which allows future developers to modify some of the parts of the tool in a simple and intuitive way, without knowing in depth the configuration of the rest of the components. In this sense, the flexibility to modify the architecture of the \textit{encoder}, the \textit{decoder} or both, without the need to intervene in the rest of the tool's \textit{pipeline} stands out. Although in this proposal the \textit{base model} of the \textit{InceptionTime} architecture has been used as the \textit{encoder}, it can be replaced by other time series deep neural architectures.

With regard to possible lines of future work, we can consider:
\begin{itemize}
    \item Improving the capabilities of the tool for multivariate time series, in order to extract valuable insights from just a subset of the variables and not necessarily use all of them.
    
    \item Expanding the input of the model with encoded information of the timestamps, such as the time of day, day of year, etc.
    
    \item Adding a new analysis mode that is fed with more than one time series at a time, and allows for a side-by-side comparison of the embeddings of each time series.
    
    \item Extending the Visual Analytics module to support 3D projection plots, as seen in analogous tools for language such as the TensorFlow embedding projector 
    \cite{smilkov_embedding_2016}.

    \item Making more experiments with different neural and non-neural based methods and compare them with the results of DeepVATS in more datasets.
\end{itemize}

\section*{Acknowledgements}
% Hodei's project, AIDA, Francesco?
This work has been funded by Grant PLEC2021-007681 (XAI-DisInfodemics) and PID2020-117263GB-100 (FightDIS) funded by MCIN/AEI/ 10.13039/ 501100011033 and, as appropriate, by “ERDF A way of making Europe”, by the “European Union” or by the “European Union NextGenerationEU/PRTR”, by European Comission under IBERIFIER - Iberian Digital Media Research and Fact-Checking Hub (2020-EU-IA-0252), by the National Science Centre, Poland under CHIST–ERA programme (NCN 2018/27/Z/ST6/03392), and by "Convenio Plurianual with the Universidad Politécnica de Madrid in the actuation line of Programa de Excelencia para el Profesorado Universitario". The authors would also like to acknowledge the Spanish State Research Agency and the European Regional Development Fund for their support through the research grant PID2020-112576GB-C22 (AEI/ERDF, UE).

\bibliography{elsarticle-template}

\pagebreak

%%
% Appendices
%% 

\appendix

\section{Synthetic datasets}
\label{appendix:synthetic_datasets}

\begin{figure}[h!]
    \centering
    \includegraphics[width=\textwidth]{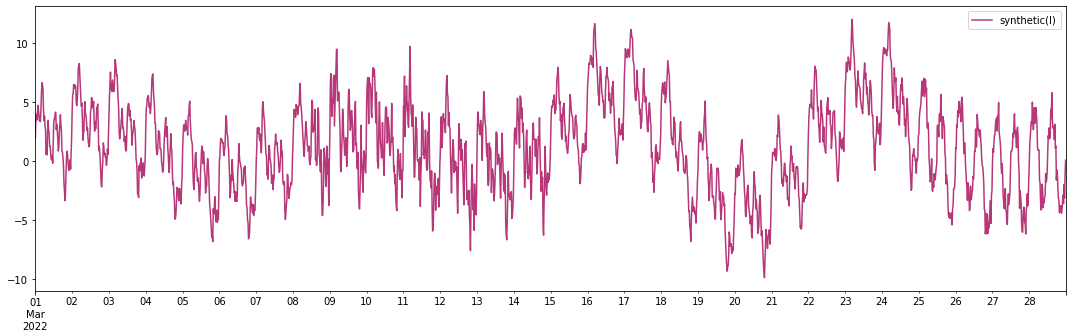}\\
    \vspace{4mm}
    \includegraphics[width=\textwidth]{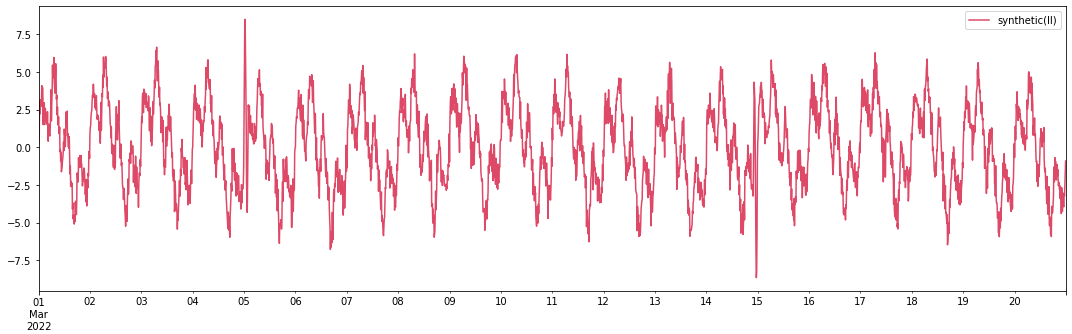}\\
    \vspace{4mm}
    \includegraphics[width=\textwidth]{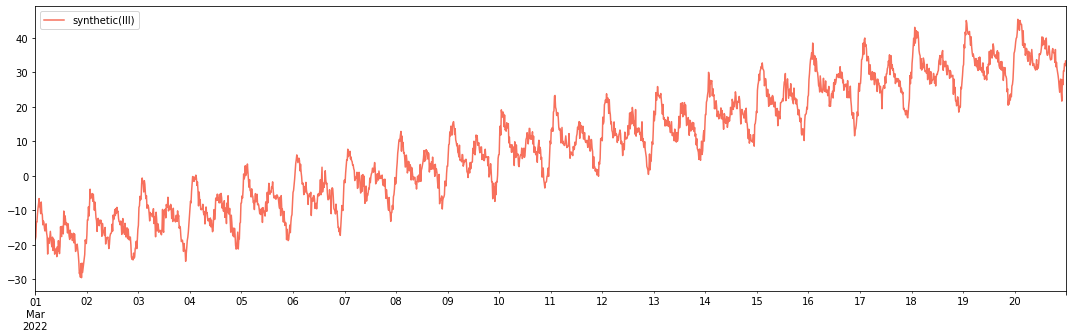}\\
    \vspace{4mm}
    \includegraphics[width=\textwidth]{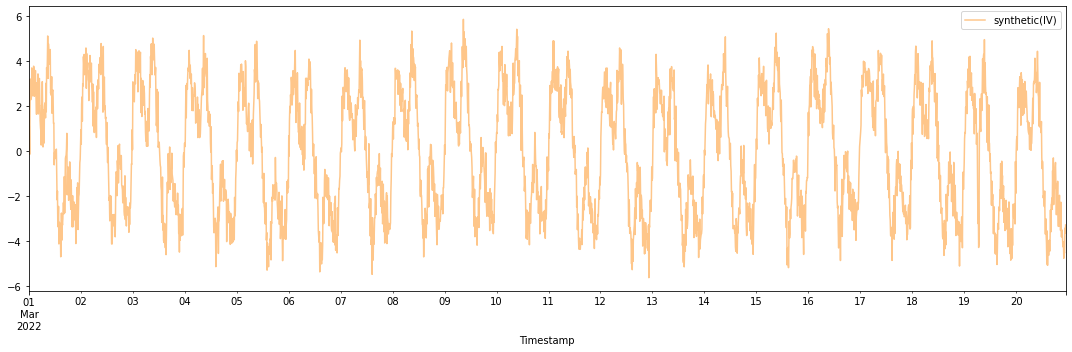}
    \caption{Synthetic time series datasets created to evaluate the capabilities of DeepVATS.}
    \label{fig:synthetic_datasets}
\end{figure}

\begin{table}[h!]
\begin{center}
\bgroup
\scriptsize
\begin{tabular}{@{}l@{}}
% \textbf{Generation formula (See Eq.  \ref{eq:gen_function})} 

\hline $\,$ \\

$g_1(t) = \left\{\begin{array}{@{}ll@{}}
f(t\mid\lambda_{2}\myeq 0.6,\lambda_{4}\myeq 1.5,\lambda_{12}\myeq 2.3,\lambda_{24}\myeq 2.3,\lambda_{168}\myeq 2.5,\gamma\myeq 1,\sigma\myeq 2) & \textrm{if } t \leq 10800\\
f(t\mid\lambda_{2}\myeq 1.5,\lambda_{4}\myeq 1.5,\lambda_{12}\myeq 2.3,\lambda_{24}\myeq 2.3,\lambda_{168}\myeq 1.5,\gamma\myeq 1,\sigma\myeq 4) & \textrm{if } 10800 < t \leq 20160\\
f(t\mid\lambda_{2}\myeq 0.6,\lambda_{4}\myeq 1.5,\lambda_{12}\myeq 2.3,\lambda_{24}\myeq 3,\lambda_{168}\myeq 5,\gamma\myeq 1,\sigma\myeq 2) & \textrm{if } 20160< t\leq 34560\\
f(t\mid\lambda_{2}\myeq 0.6,\lambda_{4}\myeq 0.5,\lambda_{12}\myeq 4.3,\lambda_{24}\myeq 1,\lambda_{168}\myeq 2,\gamma\myeq 1,\sigma\myeq 3) & \textrm{if } 34560< t\leq 40320\\
\end{array}\right.$

\\ $\,$ \\ \hline $\,$ \\

$g_2(t) = \left\{\begin{array}{@{}ll@{}}
f(\,t\mid\lambda_{2}\myeq 7,\,\lambda_{6}\myeq 2,\,\lambda_{24}\myeq 3,\,\lambda_{168}\myeq 0.5,\,\sigma\myeq 2) & \textrm{if }t \in [5760,\,5880] \cup [20040,\,20160] \\
f(\,t\mid\lambda_{3}\myeq 0.5,\,\lambda_{6}\myeq 2,\,\lambda_{24}\myeq 3,\,\lambda_{168}\myeq 0.5,\,\sigma\myeq 2) & \textrm{otherwise} \\
\end{array} \right.$

\\ $\,$ \\ \hline $\,$ \\

$g_3(t) =
0.002\cdot t + f(\,t\mid\lambda_{6}\myeq 2.5,\,\lambda_{8}\myeq 2,\,\lambda_{12}\myeq 6,\,\lambda_{24}\myeq 3,\,\lambda_{168}\myeq 1,\,\gamma\myeq -20,\,\sigma\myeq 6)$

\\ $\,$ \\ \hline $\,$ \\

$g_4(t) = \left\{ \begin{array}{@{}ll@{}}
f(\,t\mid\lambda_{4}\myeq 5,\,\lambda_{8}\myeq 2,\,\lambda_{24}\myeq 3,\,\lambda_{168}\myeq 0.5,\,\sigma\myeq 2) & \textrm{if $ 26280 < t \leq 26340$} \\
f(\,t\mid\lambda_{4}\myeq 0.5,\,\lambda_{8}\myeq 2,\,\lambda_{24}\myeq 3,\,\lambda_{168}\myeq 0.5,\,\sigma\myeq 2) & \textrm{otherwise} \\
\end{array} \right.$

\\ $\,$ \\ \hline

\end{tabular}
\egroup
\end{center}
\caption{Equations of the synthetic time series created to analyse the capabilities of DeepVATS. Equation $g_i(t)$ corresponds to the synthetic dataset $i$.}
\label{tab:table-label}
\end{table}

$\,$

\clearpage
\section{Real-world datasets}
\label{appendix:real}

\begin{figure}[h!]
    \centering
    \includegraphics[width=\textwidth]{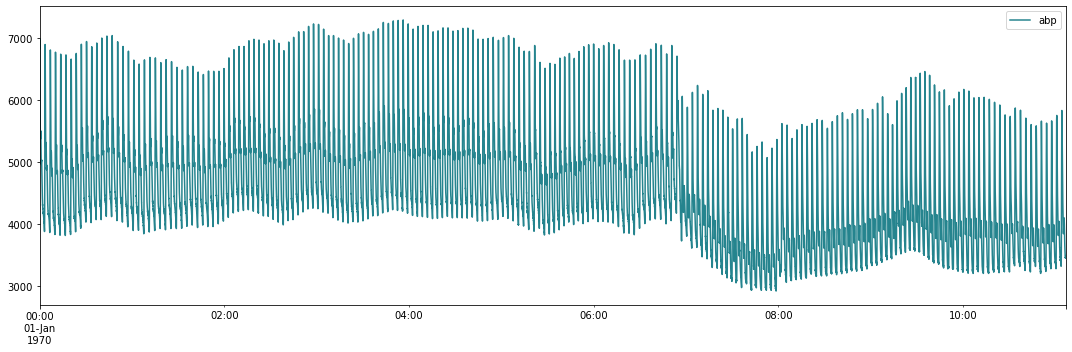}\\
    \vspace{4mm}
    \includegraphics[width=\textwidth]{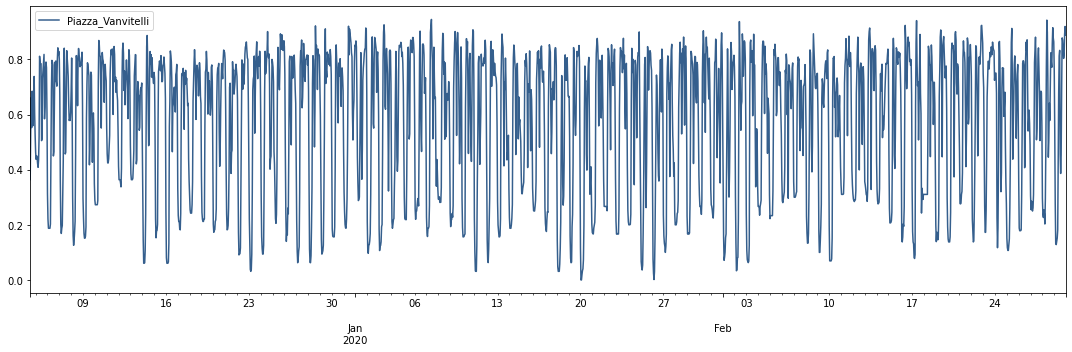}\\
    \vspace{2mm}
    \includegraphics[width=\textwidth]{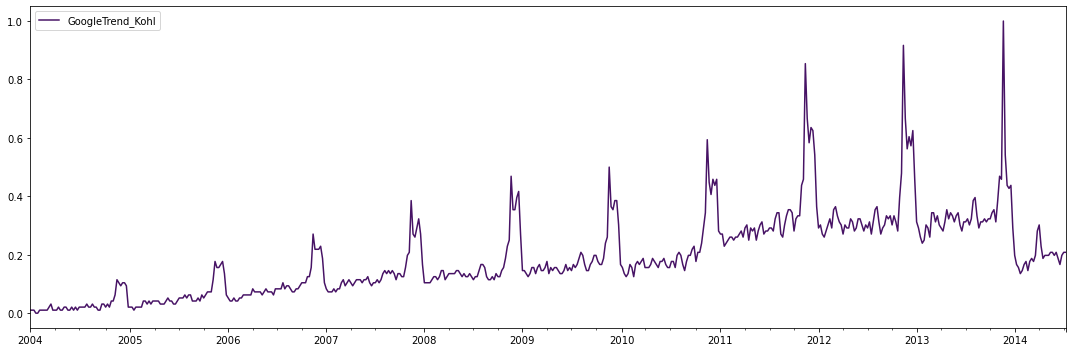}
    %\vspace{2mm}
    %\includegraphics[width=\textwidth]{figures/datasets/F10_7.png}
    \caption{Plots of time series datasets.}
    \label{fig:real_world_datasets}
\end{figure}

\clearpage
\section{Segmentation of Arterial Blood Pressure (ABP) dataset}
\label{appendix:seg}

\begin{figure}[h!]
    \centering
    \includegraphics[width=\textwidth]{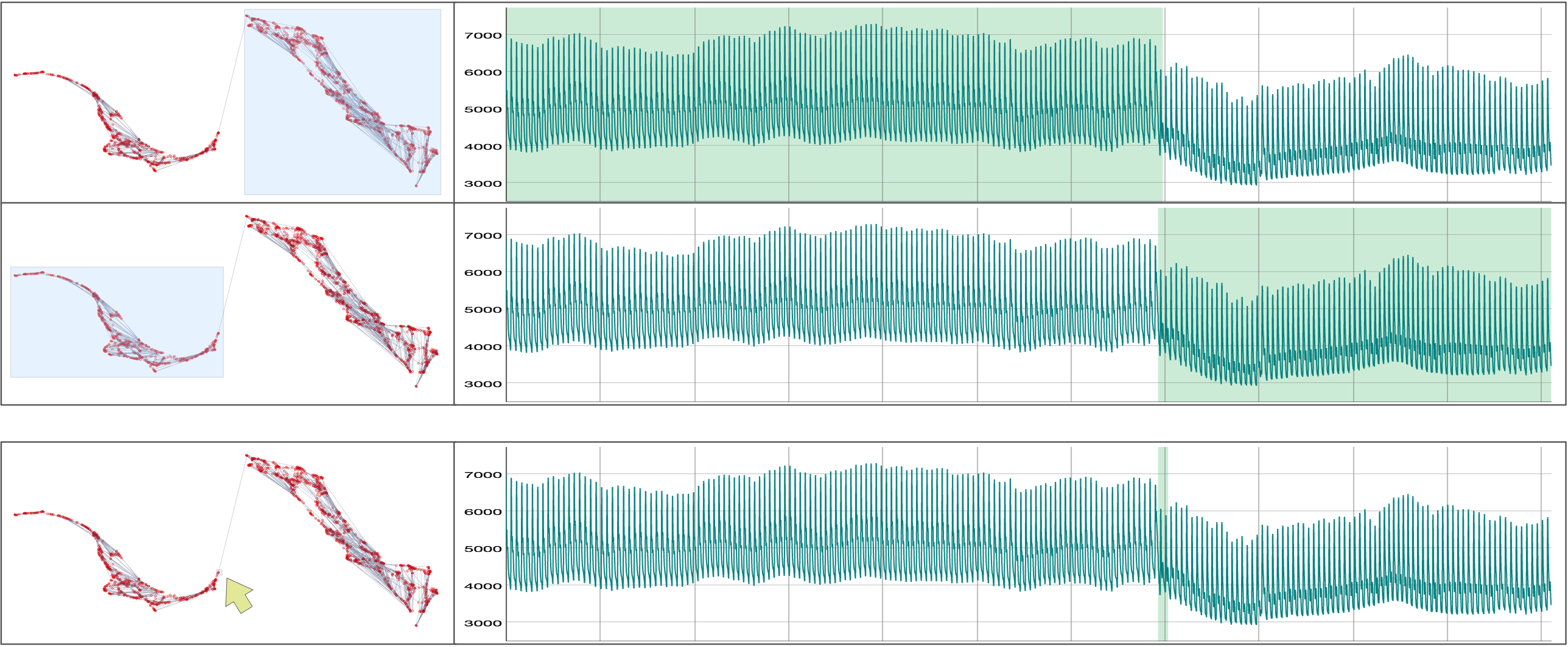}
    \caption{Segmentation and change point detection of the dataset \textit{ABP} with DeepVATS, using in the VA module, a window size of $w=210$, stride of $s=22$ and UMAP as algorithm to project the \textit{embeddings}. The first two representations show the detected segments, and the last one shows the change point between segments. The yellow arrow in the last scatter plot highlights the position of the point corresponding to the time window that contains the change point.}
    \label{fig:seg_real}
\end{figure}

After testing the behaviour of the tool in the segmentation of the \textit{S1} dataset, we proceed to analyse its use with ABP \textit{Arterial Blood Pressure} dataset. Figure~\ref{fig:seg_real} shows the result, as displayed in the VA module, of using the model trained on this series with a masking strategy analogous to that of the synthetic example (See Table \ref{table:params_data_mining}) and a variable window size of $[200, 220]$ time steps. In the the VA module, we used a window size $w=110$, a stride $s=22$ and UMAP as dimensionality reduction technique. As it can be seen, the projections plot is made up of two clearly differentiated components, joined by a grey segment. The limits of this segment are time windows containing change points.

\clearpage
\section{Repetitive patterns and outliers on the dataset \textit{S2}}
\label{appendix:motifs}

\begin{figure}[h!]
    \centering
    \includegraphics[width=\textwidth]{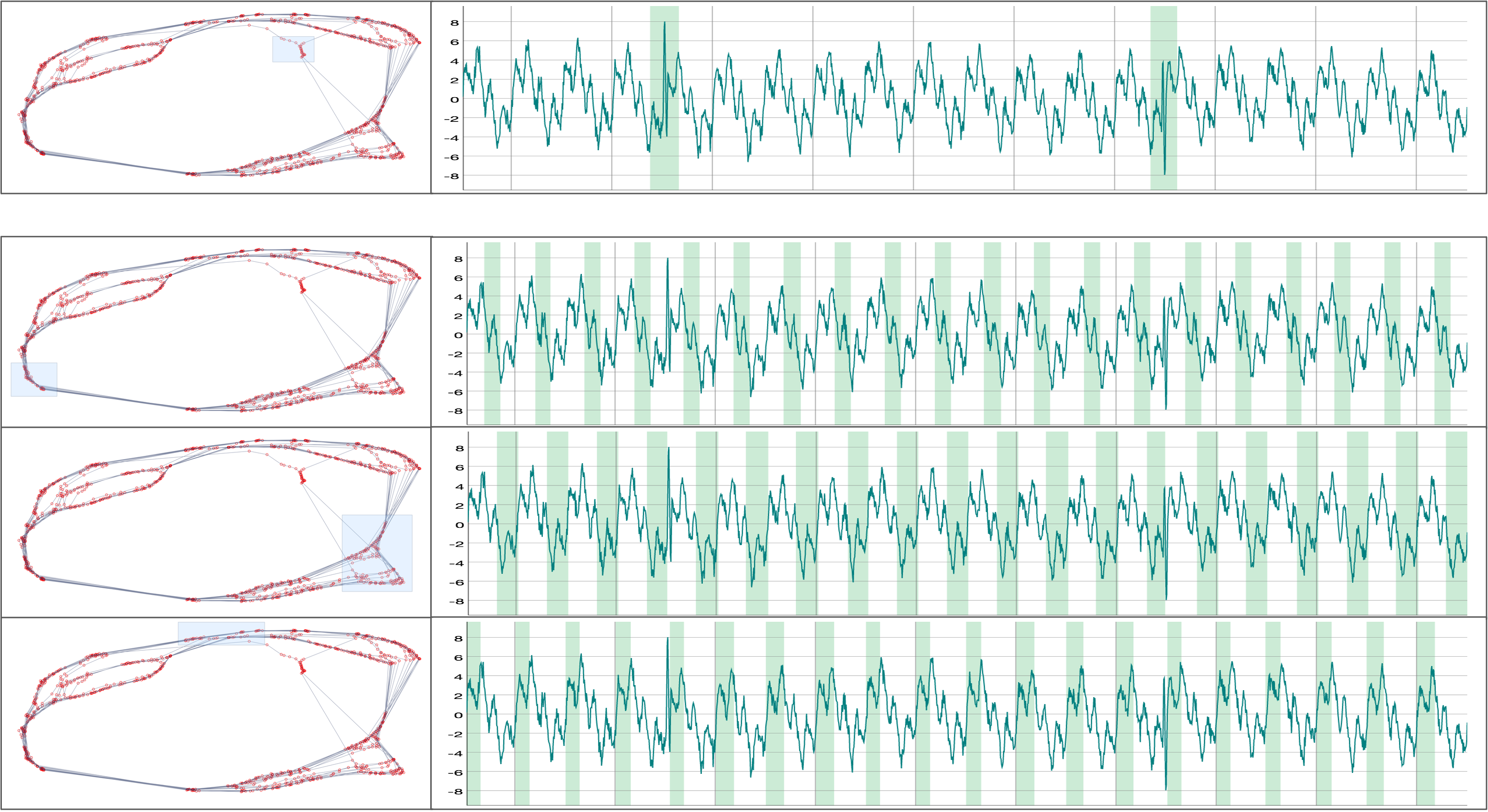}
    \caption{Anomalies (first view) and repetitive patterns (last three representations) of the synthetic dataset \textit{S2} with DeepVATS, using, in inference, a window size $w=28$, stride $s=2$ and UMAP as dimensionality reduction technique.}
    \label{fig:outliers_motifs_synthetic}
\end{figure}

As seen in Section~\ref{subsection:synthetic_data}, the dataset \textit{S2} has been constructed as the sum of components with different seasonalities, and an atypical distribution has been applied in two short intermediate periods of time. On this data, we trained the encoder model (MTSAE) with a stateless masking strategy, a masking probability of $r=0.5$ and a variable window in the range $[24,48]$. In Figure~\ref{fig:outliers_motifs_synthetic} we can see the result obtained in the VA module with a window size $w$ of $28$ samples, a \textit{stride} $s$ of $2$ timestamps, and UMAP as dimensionality reduction algorithm.

% \clearpage
% \section{Additional content for the section ``Influence of using variable window size during training"}

% \begin{table}[h!]
% \centering
% \small
% \begin{tabular}{|c|c|c|c|c|c|c|c|c|c|} 
% \hline
% \multirow{2}{*}{\textbf{Task }} & \multirow{2}{*}{\textbf{Dataset }} & \multicolumn{3}{c|}{\textbf{Masking}} & \multicolumn{3}{c|}{\textbf{Window }} & \multirow{2}{*}{\textbf{Batch}} & \multirow{2}{*}{\textbf{Epochs}} \\ 
% \cline{3-8}
%  &  & $\boldsymbol{r}$ & \textbf{stateful} & \textbf{future} & $\boldsymbol{w}$ & $\boldsymbol{w_{min}}$ & $\boldsymbol{w_{max}}$ &  &  \\ 
% \hline
% \multirow{4}{*}{\begin{tabular}[c]{@{}c@{}}Segmentation\end{tabular}} & \multirow{4}{*}{S1} & \multirow{4}{*}{$0.4$} & \multirow{4}{*}{Yes} & \multirow{4}{*}{No} & 72 & 36 & 72 & \multirow{4}{*}{16} & \multirow{4}{*}{200} \\ 
% \cline{6-8}
%  &  &  &  &  & 36 & - & - &  &  \\ 
% \cline{6-8}
%  &  &  &  &  & 54 & - & - &  &  \\ 
% \cline{6-8}
%  &  &  &  &  & 72 & - & - &  &  \\
% \hline
% \end{tabular}
% \normalsize
% \caption{Hyperparameters of the models trained to analyse te influence of using variable window size while training in the DL module}
% \label{table:params_variable_w}
% \end{table}

\end{document}